%% file: paper_MMT_R2.0.tex
\documentclass[3p,sort&compress]{elsarticle}    

\pdfoutput=1
\include{preamble_MMT}

\input{glossaries_MMT}

\usepackage{lineno} 					
\graphicspath{{Figures/}}
\begin{document}
\begin{frontmatter}
\title{Effect of boundary conditions on a high-performance isolation hexapod platform}
\author[SSC]{Alessandro Stabile}
\author[AUCK]{Vladimir V. Yotov\corref{cor1}}\ead{vladimir.yotov@auckland.ac.nz}
\author[AUCK]{Guglielmo S. Aglietti}
\author[SSC]{Pasquale De Francesco}
\author[SSTL]{Guy Richardson}

\address[SSC]{Surrey Space Centre, University of Surrey, Guildford, GU2 7XY, UK}
\address[AUCK]{Te P\={u}naha \={A}tea - Space Institute, The University of Auckland, Auckland, 1010, New Zealand}
\address[SSTL]{Surrey Satellite Technology Ltd. (SSTL), Guildford, GU2 7YE, UK}

\date{18th June 2022}
\cortext[cor1]{Corresponding author}
\journal{Mechanism and Machine Theory}

\begin{abstract}
Isolation of spacecraft microvibrations is essential for the successful deployment of instruments relying on high-precision pointing. Hexapod platforms represent a promising solution, but the difficulties associated with attaining desirable 3D dynamics within acceptable mass and complexity budgets have led to a minimal practical adoption. This paper addresses the influence of strut boundary conditions (BCs) on system-level mechanical disturbance suppression. Inherent limitations of the traditional all-rotational joint configuration are highlighted and shown to originate in link mass and rotational inertia. A pin-slider BC alternative is proposed and analytically proven to alleviate them in both 2D and 3D. The advantages of the new BC  hold for arbitrary parallel manipulators and are demonstrated for several hexapod geometries through numerical tests. A configuration with favourable performance is suggested. Finally, a novel planar joint that allows the physical realisation of the proposed BC is described and validated. Consequently, this work enables the development of platforms for microvibration attenuation that do not require active control.
\end{abstract}

\end{frontmatter}

\section{Introduction}

An ultra-quiet environment is indispensable for space systems employing high-precision instruments like optical sensors, space interferometers and laser communication equipment. Maintaining the necessary stability can be compromised by low-level vibrations produced by multiple onboard mechanisms. The most prominent sources are rotating devices, namely momentum and reaction wheel assemblies \cite{richardson:rw_test_1, mv_zhang_11, mv_li_18}, together with cryocoolers \cite{mv_carte_21}. Such disturbances propagating through the satellite structure are collectively referred to as microvibrations. They normally occur in the form of micro-g perturbations whose practically significant frequency content spans the \SI{1}{\Hz} to \SI{1}{\kHz} range. Minimisation of the induced detrimental effects has been the subject of ongoing academic and industrial efforts. Nonetheless, continuing advancements in the field have been matched by the increasingly stringent stability demands of present and upcoming missions. For comprehensive reviews of the state of the art, the reader is referred to \cite{mv_komatsu_14, liu:recent_advances_microvib_isol, mv_li_21}. 
	
Tasks contingent on accurate positioning have motivated the rapid development of parallel manipulators, which have several beneficial characteristics \cite{patel:hex_survey, dasgupta:hex_review}. While generally lacking the dexterity and large workspace of their serial counterparts, they are capable of sustaining greater loads, have lower inertia, higher structural stiffness and some built-in redundancy. More importantly, they are amenable to simpler control and pose reduced susceptibility to certain errors. The hexapod platform is among the most popular and replicated configurations. It comprises six variable-length struts arranged in a way as to allow motions in six \glspl{acr:dof} by using universal and spherical joint connections. It was first built in the early 1950s \cite{gough:tyre_test}, but only became publicly known and received the attention of the scientific community a decade later. Its subsequent diffusion was due to the nearly simultaneous publications by Stewart \cite{stewart:platform}, Gough \textit{}\cite{gough:tyre_test_2} and Cappel \cite{cappel:platform}. Nowadays hexapods find regular use in terrestrial applications such as flight simulators \cite{dong:flight_sim}, fast pick-and-place robots \cite{connolly:pick_robot1,angel:pick_robot2}, micro-surgical procedures \cite{kumar:surgery1, meggiolaro:surgery2} and machine tooling \cite{karimi:hex_machine_tool, ren:hex_machine_tool_2, chiu:hex_machine_tool_3}.

Within the space hardware domain, parallel manipulators have been identified as a possible means of meeting the aforesaid in-orbit stability requirements \cite{mv_qin_20}. The attempts to decouple sensitive payloads from the noisy mechanical environment have resulted in the creation of various hexapod systems. Examples include the vibration isolation and suppression system (VISS) which had passive viscous dampers \cite{cobb:viss}, the satellite ultraquiet isolation technology experiment (SUITE) made of active struts with embedded piezoelectric actuators \cite{anderson:suite} and the miniature vibration isolation system (MVIS), which combined passive and active approaches into a hybrid solution \cite{jacobs:mvis_II_application}. Regardless, these mechanisms rarely see space mission integration due to persistent limitations that are yet to be addressed. In particular, the dynamic complexity \cite{mv_wu_17}, considerable amount of added mass and need for control algorithms and sensors \cite{liu:recent_advances_microvib_isol} are usually not counterbalanced by a sufficient attenuation enhancement. To that end, research has largely been dedicated to the improvement of strut isolation performance through the development of novel damping methods \cite{preumont:single_stage_strut, lee:hybrid_isol_vca, stabile:1_emsd, stabile:sms, stabile:strut_concept}. However, overall vibration transmissibility is dictated to a comparable extent by factors such as system-level topology and dynamic interaction between all of its constitutive elements \cite{mv_wu_17, mv_yang_19}.

Recent advances pertaining to architectural and control optimisation have brought significant analytical rigour into the hexapod platform design process. Configuration-specific techniques can be traced back to the classic cubic section geometry \cite{mv_geng_94}. Increasingly sophisticated models have ensued, e.g. with the current incorporation of nonlinear control and parameter uncertainties \cite{mv_wu_15}, flexure leg joints \cite{mv_yang_19} and, for passive systems, negative stiffness magnetic springs \cite{mv_zheng_18, mv_wang_20}. Contemporary works often explore the conditions necessary for the realisation of dynamic isotropy. In other words, their objective is to find parameter sets that result in equal \glspl{acr:nf} across the six fundamental rotational and translational modes of the primary mass. Dynamic isotropy is appealing due to the reduced positioning errors and simplified control algorithms in the case of active struts \cite{yang:hex_dynamic_isotropy, jiang:hex_dyn_isotropy, kourosh:hex_dyn_isotropy2, afzali:dynam_isotropy, afzali:dynam_isotropy2}. The main approaches taken in geometrical optimisation can be classified into structure-oriented and Jacobian-oriented, further elaborated in \cite{jiang:hex_dyn_isotropy}. A simplifying assumption commonly found in literature is that strut mass can be ignored. This is deemed acceptable, since the combination of large payload-to-link mass ratio, small displacements and low frequency range renders strut inertia effects negligible. However, it has recently been shown that the influence of link mass on the global dynamics becomes dominant at higher frequencies \cite{wu:hex_dyn_isotropy_strut_mass, afzali:strut_inertia}.

\begin{figure}[htb]
\centering
	\pgfplotsset{every axis/.append style=log subplots,} 
	\begin{tikzpicture}
		\begin{axis}[ymin=-85,ymax=50,xmin=0.1,xlabel={Frequency (Hz)},ylabel={Magnitude (dB)}]
		\addplot[rc,thin] table {TikzData/F1-hexa1.tsv};
			\addlegendentry{Hexapod $1$}
		\addplot[thick,dotted] table {TikzData/F1-hexa2.tsv};
		 	\addlegendentry{Hexapod $2$}
		\addplot[bc,thin] table {TikzData/F1-hexa3.tsv};
		 	\addlegendentry{Hexapod $3$}
		\end{axis}
	\end{tikzpicture}
\caption{Vertical input-vertical output force TF for hexapods with identical struts and payload but different geometry}
\label{fig: plateau for different hexapods}	
\end{figure}
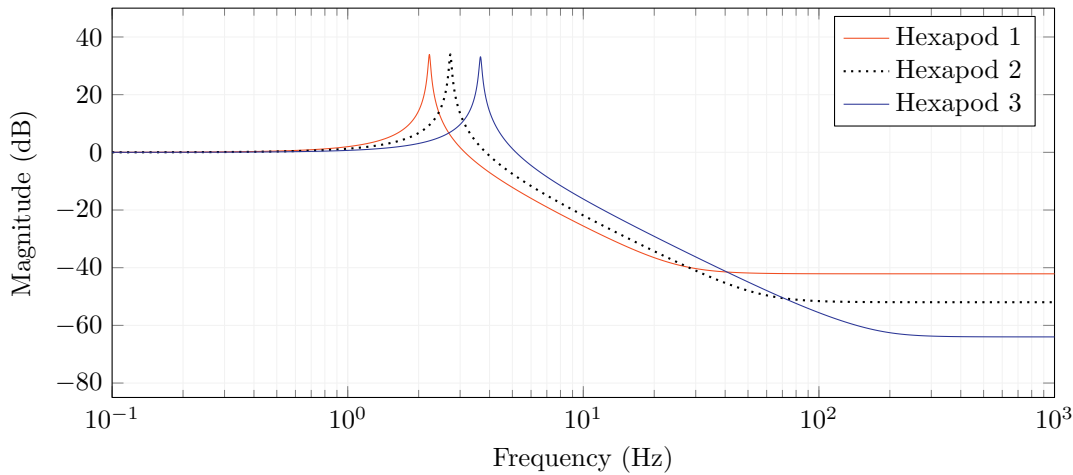

Typical \glspl{acr:tf} obtained by modifying hexapod geometry, but not payload and strut specifications, are reported in \Cref{fig: plateau for different hexapods}. A formal parametrisation of hexapod geometry is given in \Cref{ssec: hexapod geometry parametrisation}. Here, stiffness, damping, masses and top platform radius were kept constant. On the other hand, the angle between link pairs, positions of joints to the moving platform, as well as link-to-payload angles were allowed to vary. Altering the aforesaid quantities produces slightly different TFs, as seen in \Cref{fig: plateau for different hexapods}, but the overall disturbance transmission properties remain unchanged. In particular, one immediately infers from the plot that instead of maintaining the expected slope of \SI{-40}{\dB\per\dec}, the curves plateau after the resonance, which arrests the attenuation at a constant level. In fact, this behaviour closely resembles the findings in \cite{wu:hex_dyn_isotropy_strut_mass}. This effect is indeed attributed to the combination of link inertia and the use of all-pin \gls{acr:bc}, as explained in \Cref{sec: all-rotational limitations}. Given a fixed strut design, payload and BCs, the plateau cannot be avoided by changing the geometrical configuration.

This paper prescribes a novel approach towards the development of platforms for microvibration mitigation without active control. They are preferable for space missions from a cost, complexity and robustness perspective, but current systems have intrinsic limitations. Indeed, a principal constraint to reducing mid- to high-frequency transmissibility is identified in \Cref{sec: all-rotational limitations}. It is proven that the standard pin-pin link \gls{acr:bc}, originating in the exclusive use of rotational joints, inevitably produces a nonzero TF lower bound. A pin-slider alternative that has not previously been considered in literature is proposed in \Cref{sec: alternative BCs}. It is capable of restoring force TFs to theoretically ideal ones ascribed to massless links. Equations of motion for the new BC, encompassing arbitrary link number and orientation, are derived in \Cref{sec: pin-slider BC}. Analytical models are verified in 2D and 3D, corroborating the benefits of the proposed BC. Hexapod geometry recommendations are given based on results from the analytical model. Finally, in \Cref{sec: planar joint}, a novel planar joint design is demonstrated and tested as a potential implementation of the slider BC.

\section{Limitations of all-rotational joint systems}
\label{sec: all-rotational limitations}
\subsection{Paper notation}\label{ssec: paper notation}

The notation style employed throughout the article is first established. Matrices and vectors are respectively indicated by upper and lower case bold letters, e.g. $\m{M}$, $\vc{f}$. Their components are indexed as usual, that is $\vc{f} = [\begin{matrix} F_1, F_2, \dots \end{matrix}]\tr{}$. All scalar quantities are in regular type face, with the ones pertaining to force, torque and moment in upper case. Dot denotes time derivative, whereas bar and hat accents are respectively used for intermediate variables and specification of payload position and orientation, e.g. $\vc{\bar{\uptau}}$ and $\m{\hat{R}}$, $\vc{\hat{d}}$. The skew-symmetric and $\diag$ operators, defined in \Cref{ssec: equations of motion}, are used for matrix construction.

\subsection{Massless strut idealisation}\label{ssec: massless struts}

A hexapod in its standard initial position is an isostatic structure \cite{guest:hexapod_isostatic}, that is, both statically and kinematically determinate. This condition is achieved if all struts are assumed infinitely rigid and connected to the upper and lower platforms using pin joints, i.e. universal, spherical and their flexure equivalents. Normally, this is facilitated by S-joints for the fixed base and U-joints for the equipment. The importance of this feature is that manufacturing and assembling tolerances and errors can be dealt with without the introduction of extra deformations or stresses in the system. Given the reasonable assumption that each link is a single-DOF device and that the payload is a rigid body, the mobile platform exhibits the anticipated six spatial motion DOFs. 

\begin{figure}[hbt]
\centering
	\captionsetup[subfigure]{skip=\subfigskip}
	\setl{figh}{0.335\columnwidth}
\subcaptionbox{
	\label{fig: ideal_s (a)}}{
	\begin{tikzpicture}
	\node[inner sep=0pt, anchor=south west] (img) at (0,0) {\includegraphics[height=1.0556\figh]{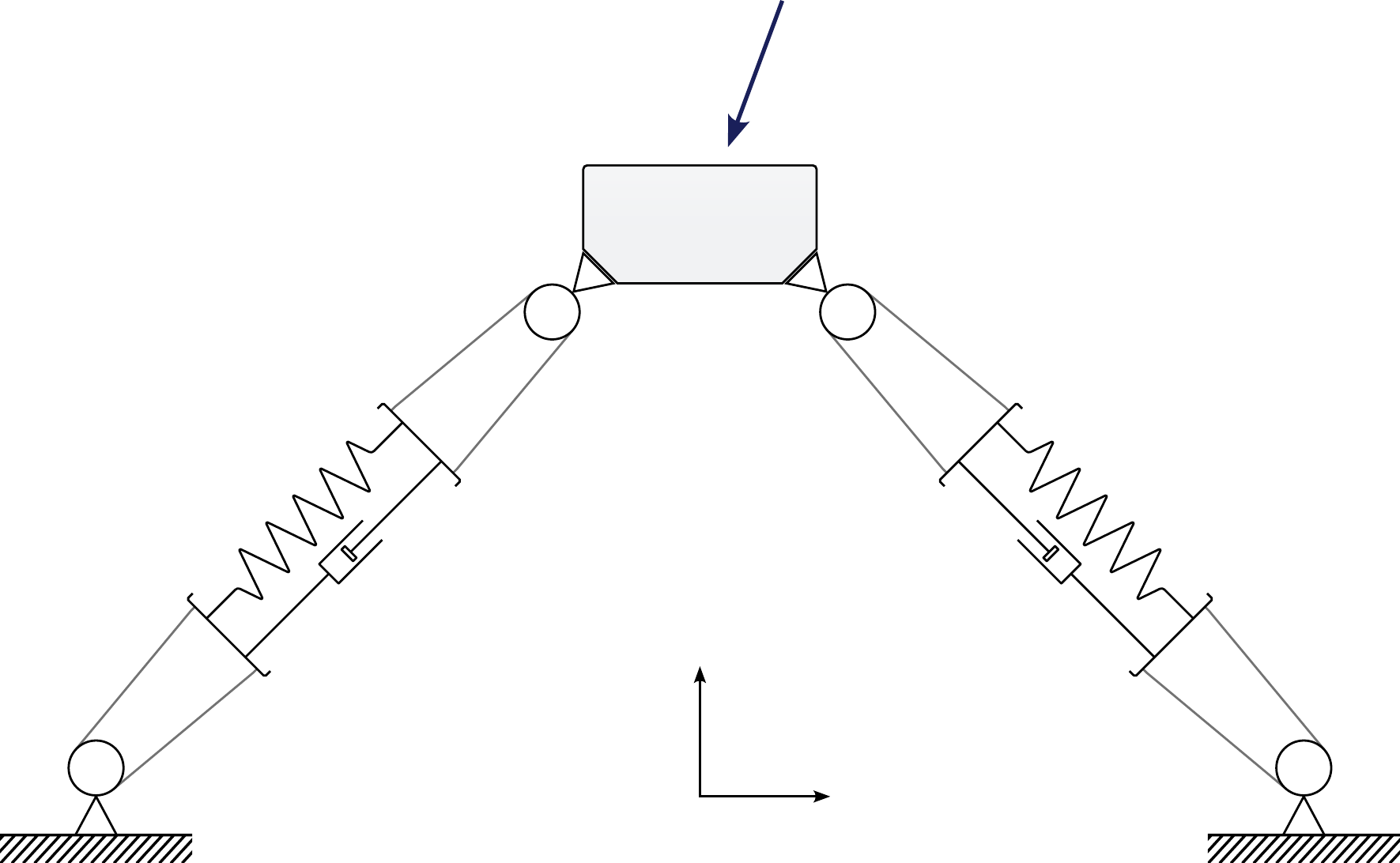}};
	\begin{scope}[ x={($0.01*(img.south east)$)}, y={($0.01*(img.north west)$)}, ]
		\begin{pgfonlayer}{annotation}
			\node[inner sep=0pt,anchor=south] at (50,24) {$z$};
			\node[inner sep=0pt,anchor=west] at (60,8) {$x$};
			\node[inner sep=0pt,anchor=north west] at (55.0,93.1) {$F_\text{in}$};
			\node[inner sep=-5pt,anchor=south] at (50.0,73.0) {$m_p$};
		 	\node[inner sep=1pt,anchor=north west] at (26.5,32.5) {${c}$};
	 		\node[inner sep=1pt,anchor=north east] at (73.5,32.5) {${c}$};	  		
	 		\node[inner sep=0pt,anchor=south east] at (19.5,43.5) {${k}$};
	 		\node[inner sep=0pt,anchor=south west] at (80.5,43.5) {${k}$};
		\end{pgfonlayer}
	\end{scope}
\end{tikzpicture}}
\hspace{0.03\columnwidth}
\subcaptionbox{
	\label{fig: ideal_s (b)}}{
	\begin{tikzpicture}
	\node[inner sep=0pt, anchor=south west] (img) at (0,0) {\includegraphics[height=1.0556\figh]{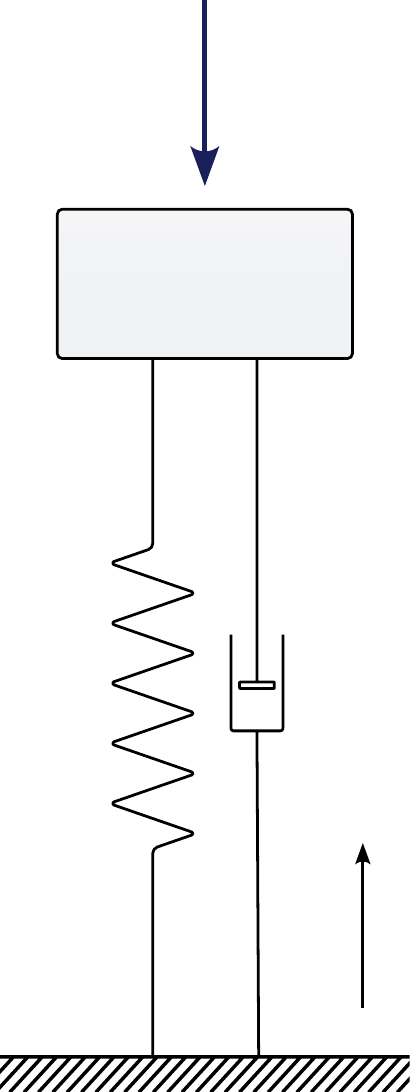}};
	\begin{scope}[ x={($0.01*(img.south east)$)}, y={($0.01*(img.north west)$)}, ]
		\begin{pgfonlayer}{annotation}
			\node[inner sep=0pt,anchor=south] at (88.3,24) {$z$};
			\node[inner sep=0pt,anchor=north west] at (54.7,93.1) {$F_\text{in}$};
			\node[inner sep=-5pt,anchor=south] at (50.0,73.0) {$m_p$};
			\node[inner sep=0pt,anchor=south] at (83.0,35.0) {$\bar{c}$};	  		
			\node[inner sep=0pt,anchor=south] at (14.0,35.0) {$\bar{k}$};
		\end{pgfonlayer}
	\end{scope}
\end{tikzpicture}}
\caption{Idealised bipod diagram: \subref{fig: ideal_s (a)} massless struts, \subref{fig: ideal_s (b)} equivalent representation as a 1-DOF oscillator}
\label{fig: ideal_s}
\end{figure}
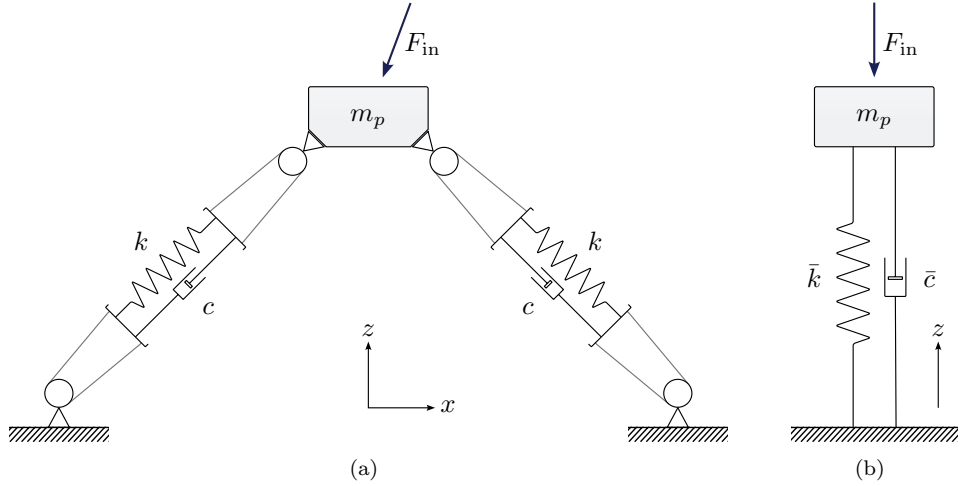

Radially symmetric hexapods are prevalent in practice. They are composed of three planar two-chain structures, referred to as bipods in this paper. In a 2D setting, bipods are isostatic and many of their dynamic characteristics directly carry over to 3D. Therefore in this section, the inadequacy of pin-pin link BCs for mid- to high-frequency passive isolation is explored within a planar analysis framework. Considering bipods as the principal building blocks of a parallel manipulator also presents an opportunity to formulate slightly more compact 3D equations of motion, as demonstrated in \Cref{sec: pin-slider BC}. When strut mass is ignored, a pin-pin bipod simplifies to the schematic reported in \Cref{fig: ideal_s (a)}. It is noted that in general, the presence of a damping element, such as a dashpot, does not affect the conclusions drawn throughout this paper. Nevertheless, viscous damping is included in all subsequent derivations for the sake of completeness.

Upon a closer examination, it is evident that with massless, one-dimensional ideal struts, the system's dynamics are identical to a 1-DOF oscillator, e.g. \Cref{fig: ideal_s (b)}. Equivalent stiffness $\bar{k}$ and damping $\bar{c}$ are easily calculated from the initial planar angle between the struts and the input force angle. When the two links are orthogonal, the force transfer function for aligned input-output becomes independent of excitation direction. This is established in \Cref{ssec: real struts}, and implies dynamic isotropy in the structure's plane. The aforesaid TF is characterised by a single resonance due to a pair of complex conjugate poles, followed by a roll-off slope of \SI{-40}{\dB\per\dec}, as depicted in \Cref{fig: bipod shear}. In the presence of viscous damping, the slope reduces to \SI{-20}{\dB\per\dec}, hence the transfer function limit at infinity remains zero. These basic observations suggest that the force transmissibility of the structure considered strictly decreases with frequency past the resonance. However, this is not the case once the effects of link mass are accounted for, as conferred next.

\subsection{Real strut bipods}\label{ssec: real struts}

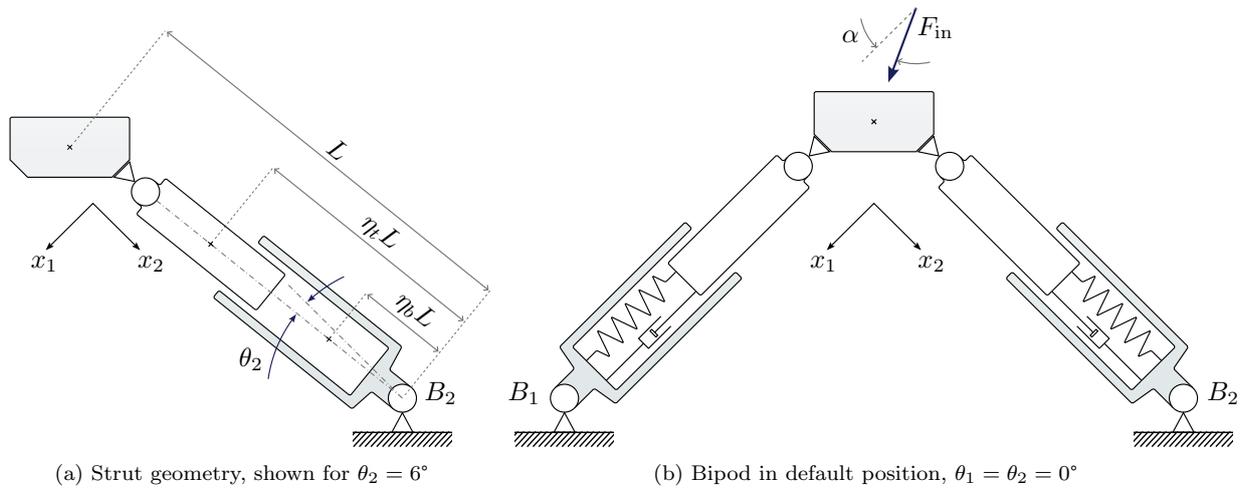
\begin{figure}[htb]
\centering
	\captionsetup[subfigure]{skip=\subfigskip}
	\setl{figh}{0.335\columnwidth}
\subcaptionbox{
	{Strut geometry, shown for $\theta_2=\ang{6}$}
	\label{fig: 2D pin-pin bipod (a)}}{
	\begin{tikzpicture}
	\node[inner sep=0pt, anchor=south west] (img) at (0,0) {\includegraphics[height=\figh]{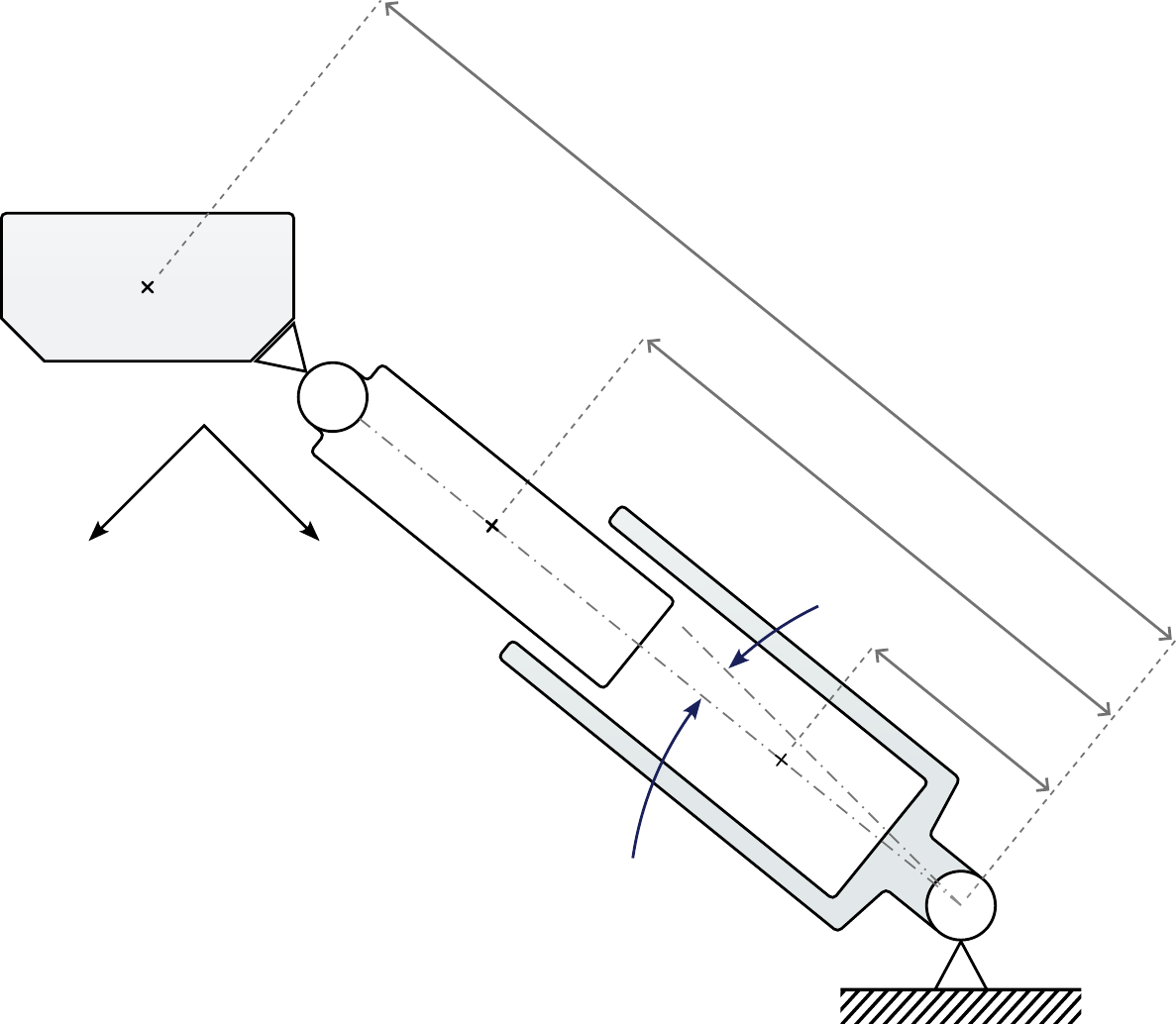}};
	\begin{scope}[ x={($0.01*(img.south east)$)}, y={($0.01*(img.north west)$)}, ]
		\begin{pgfonlayer}{annotation}
 			\coordinate (C) at (82,29.5);
 				\node[inner sep=2pt,anchor=south,rotate around={-39:(C)}] at (C) {$\eta_b L$};
 			\coordinate (D) at (74.8,48.3);
				\node[inner sep=2pt,anchor=south,rotate around={-39:(D)}] at (D) {$\eta_t L$};
 			\coordinate (E) at (66.3,68.5);
				\node[inner sep=2pt,anchor=south,rotate around={-39:(E)}] at (E) {$L$};
 			\foreach \Point in {
 		 				 		}{\node at \Point {$\circ$};}
  		 	\node[inner sep=0pt,anchor=south west] at (86.2,10.05) (B2) {$B_2$};
 			\node[inner sep=0pt,anchor=north east] at (10.3,45.91) {$x_1$};
 			\node[inner sep=0pt,anchor=north west] at (26.6,45.91) {$x_2$};
 			\node[inner sep=1pt,anchor=south east] at (53.5,17.5) {$\theta_2$};
		\end{pgfonlayer}
	\end{scope}	
	\end{tikzpicture}}
\subcaptionbox{
	{Bipod in default position, $\theta_1=\theta_2=\ang{0}$}
	\label{fig: 2D pin-pin bipod (b)}}{	
	\begin{tikzpicture}
		\node[inner sep=0pt, anchor=south west] (img) at (0,0) {\includegraphics[height=1.0556\figh]{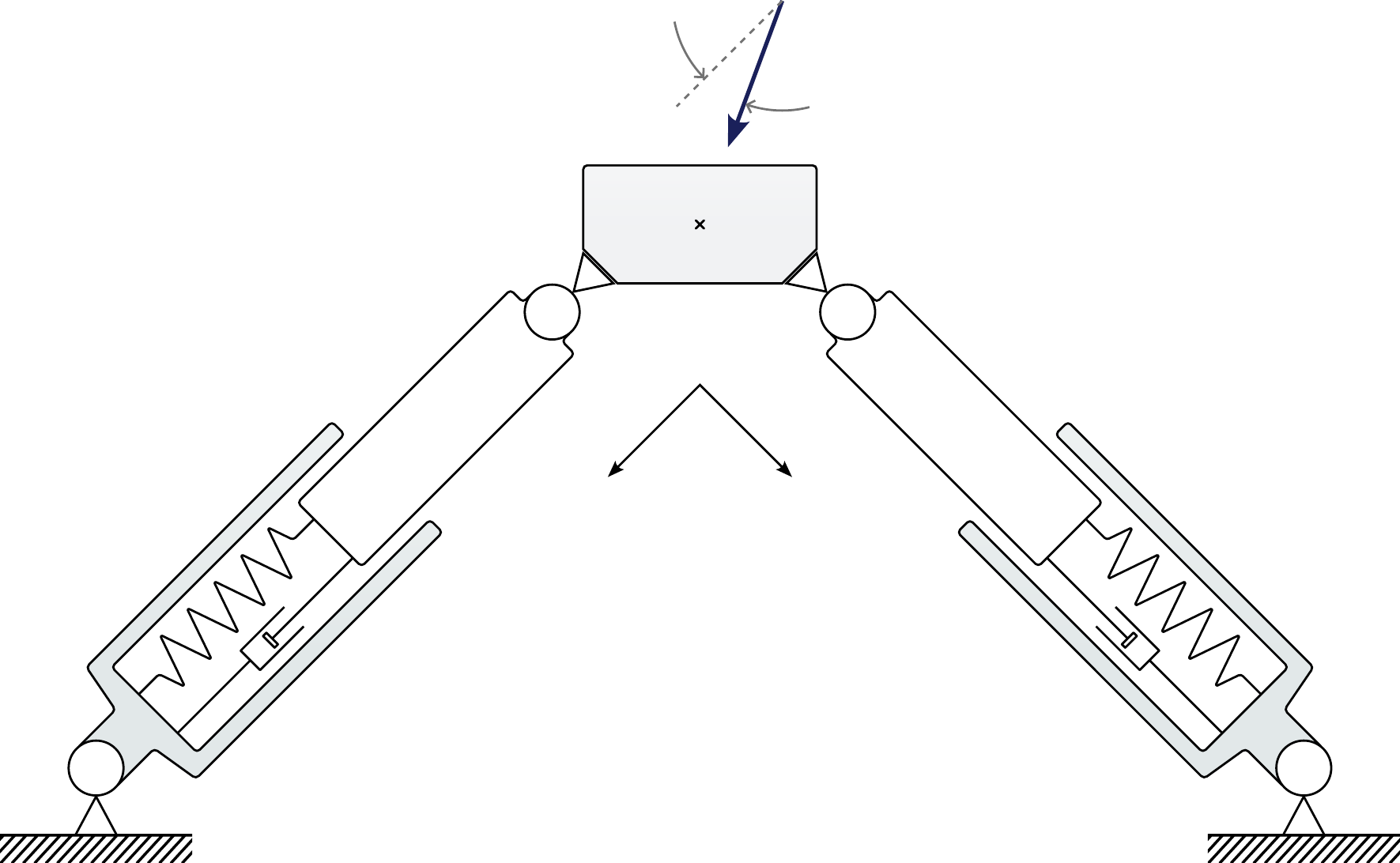}};
	 	\begin{scope}[ x={($0.01*(img.south east)$)}, y={($0.01*(img.north west)$)}, ]
 		 	\begin{pgfonlayer}{annotation}
 		 		\node[inner sep=0pt,anchor=south west] at (-1,9.5) (B1) {$B_1$};
 		 		\node[inner sep=0pt,anchor=south east] at (101,9.5) (B2) {$B_2$};
 		 	 	\node[inner sep=0pt,anchor=north east] at (45,43.5) {$x_1$};
 				\node[inner sep=0pt,anchor=north west] at (56.1,43.5) {$x_2$};
 		 		\node[inner sep=0pt,anchor=north west] at (56,98) {$F_\text{in}$};
 		 		\node[inner sep=0pt,anchor=north west] at (45.5,94.7) {$\alpha$};
 			\end{pgfonlayer}
 		\end{scope}
	\end{tikzpicture}}
\caption{Diagram of a bipod with struts made of two rigid components, each of which has a nonzero mass}
\label{fig: 2D pin-pin bipod}
\end{figure}

In all-rotational joint manipulators, link mass of less than 5\% of the primary mass is sufficient to severely degrade the system's mid- to high-frequency vibration attenuation. To illustrate the concept, a bipod with perpendicular struts, as depicted in \Cref{fig: 2D pin-pin bipod}, is studied hereafter. Each link has a stiffness ${k}$, damping ${c}$, nominal length $L$ and a mass $m_s = m_t + m_b$, where $m_t$ and $m_b$ respectively refer to the components attached to the mobile and fixed platforms. The former has inertia $I_t$ and its \gls{acr:cm} lies at a distance $\eta_t L$ from its base attachment point. Similarly, $I_b$ and $\eta_b L$ describe the same quantities for the bottom part. Note that $\eta_t$ is assumed to be constant in light of the small system workspace in microvibration applications. The payload is characterised by parameters $m_p$ and $I_p$. Subscripts $1$ and $2$ indicate the left- and right-hand side struts, respectively, whereby $\theta_i$ is the rotation of strut $i$ about $B_i$.

Let $\alpha$ denote the angle between a force applied to the payload and the $x_1$-axis, as shown in \Cref{fig: 2D pin-pin bipod (b)}. Consider an $x_1$-aligned excitation, i.e. $\alpha=0$. The usual small angle approximations result in zero translations along $x_2$ for any \gls{acr:cm}. Examining the bipod's total linear momentum as well as its angular momentum with respect to $B_2$,
\begin{subequations}\label{eq: 2D momentum}
 \begin{align}
 F_1 + F_2 + F_\text{in} &= m_\text{dyn} \ddot{x}_1 							\label{eq: 2D momentum (a) lin} \\
 L(F_1 + F_\text{in}) 	 &= I_s \ddot{\theta}_2 + L(m_p + m_t) \ddot{x}_1		\label{eq: 2D momentum (b) ang}
 \end{align}
\end{subequations}
Here, $F_i$ is the reaction force at $B_i$, $I_s$ the rotational inertia of the right link about $B_2$ and $L(m_p + m_t)\ddot{x}_1$ arises due to the purely linear motion of the payload and left strut's upper piece. Utilising $\ddot{x}_1 \approx L \ddot{\theta}_2$ and substituting an expression for $\ddot{x}_1$ obtained from \eqref{eq: 2D momentum (b) ang} into \eqref{eq: 2D momentum (a) lin},
\begin{subequations}\label{eq: 2D F1,F2,beta}
 \begin{align} 
  F_1 	&= -{k} x_1 -{c}\dot{x}_1 										\label{eq: 2D F1,F2,beta (a) axial}\\
  F_2  	&= \lambda (F_1 + F_\text{in}) 									\label{eq: 2D F1,F2,beta (b) shear}
 \end{align}
\end{subequations}
The system's dynamics are fully determined by ${c}$, ${k}$ and the coefficients
\begin{subequations}
 \begin{align}\label{eq: 2D mdyn,Is,beta}
  m_\text{dyn} &= m_p + m_t + \eta_t m_t + \eta_b m_b 					\\
  I_s &= I_t + I_b + (m_t\eta_t^2  + m_b\eta_b^2)L^2 					\\
  \lambda &= \frac{m_\text{dyn} L^2}{I_s + (m_p + m_t) L^2} - 1
 \end{align}
\end{subequations}
In particular, the axial reaction $F_1$ and shear reaction $F_2$ can be analysed in the frequency domain by taking Laplace transforms of \eqref{eq: 2D momentum (a) lin} and \eqref{eq: 2D F1,F2,beta},
\begin{subequations}\label{eq: 2D TFs}
 \begin{align}
  \text{TF}_\text{axial} &= \frac{-(\lambda+1)({c} s + {k})}{m_\text{dyn}s^2 + (\lambda+1)({c} s + {k})} 	\label{eq: 2D TFs (a) axial} \\
  \text{TF}_\text{shear} &= \frac{\lambda m_\text{dyn} s^2}{m_\text{dyn}s^2 + (\lambda+1)({c} s + {k})} 	\label{eq: 2D TFs (b) shear}
 \end{align}
\end{subequations}
where $s = \sigma+j\omega$ for $\sigma\in\mathbb{R}$, circular frequency $\omega$ and $j=\sqrt{-1}$. 
The initial choice of input angle $\alpha=0$ intends solely to demonstrate the different response behaviour in the longitudinal and transverse directions. To obtain the general TFs, allow $\alpha\neq 0$. The $[\begin{matrix} x_1 & x_2 \end{matrix}]\tr{}$ base joint TF vectors read
\begin{subequations}\label{eq: 2D general TFs}
 \begin{align}
  \text{TF}_{B_1} & = 
   \tr{ \begin{bmatrix}  \text{TF}_\text{axial} \cos\alpha  & \text{TF}_\text{shear} \sin\alpha  \end{bmatrix} }  \\
  \text{TF}_{B_2} & = 
   \tr{ \begin{bmatrix}  \text{TF}_\text{shear} \cos\alpha  & \text{TF}_\text{axial} \sin\alpha  \end{bmatrix} }
 \end{align}
\end{subequations}
The idealised case discussed in \Cref{ssec: massless struts} corresponds to $\text{TF}_\text{shear} = 0$ in \eqref{eq: 2D general TFs}. An elementary verification with $m_t = m_b = I_t = I_b = 0$ reveals that indeed $\lambda = 0$ for massless struts. 

It is noted that both TFs in \eqref{eq: 2D TFs} are characterised by the same couple of complex conjugate poles. However, \eqref{eq: 2D TFs (b) shear} presents an additional zero at the origin (two in the absence of damping). This combination of poles and zeros results in a Bode diagram with an initial slope of +\SI{40}{\dB\per\dec}. After the resonance, the plot reaches a plateau because the poles are responsible for a change of slope of \SI{-40}{\dB\per\dec}. Calculating the level of the plateau from \eqref{eq: 2D general TFs} is trivial, since
\begin{equation}
 \smashoperator{\lim_{\omega\to\infty}} \text{TF}_\text{shear} = \lambda		,\quad\;
 \smashoperator{\lim_{\omega\to\infty}} \text{TF}_\text{axial} = 0 
\end{equation}

Interestingly, designing a bipod for $\lambda=0$ is possible, but it still does not rectify the all-rotational joint BC's shortcomings when spatial effects are involved. Consider any out-of-plane motion of $P_1$. The transverse reactions at $B_1$ and $B_2$ are inevitably proportional to a weighted sum of masses akin to $m_\text{dyn}$. Barring joint friction, the induced TFs are nonzero constants.

\begin{figure}[htb]
\centering
	\pgfplotsset{every axis/.append style=log subplots,} 
\begin{tikzpicture}
	\begin{axis}[ymin=-120,ymax=60,xmin=0.1,xlabel={Frequency (Hz)},]
	\addplot[dotted] table {TikzData/F-F dyn iso massless.tsv};
		\addlegendentry{$F\shortminus F$, idealised}
	\addplot[bc,thin] table {TikzData/axialTF.tsv};
	 	\addlegendentry{$F_\text{in}\shortminus F_\text{axial}$}
	\addplot[rc,thin] table {TikzData/shearTF.tsv};
	 	\addlegendentry{$F_\text{in}\shortminus F_\text{shear}$}
	\addplot[dbc,very thick,dashed] table {TikzData/totalTF.tsv};
	 	\addlegendentry{$F_\text{in}\shortminus F_\text{total}$}
	\end{axis}
\end{tikzpicture}
\caption{Shear, axial and total transmitted forces due to an $x_1$ excitation ($\alpha=0$) for the real strut case depicted in \Cref{fig: 2D pin-pin bipod}. An undamped, massless link idealisation TF of the same bipod is superimposed for comparison (dotted line)}
\label{fig: bipod shear}	
\end{figure}
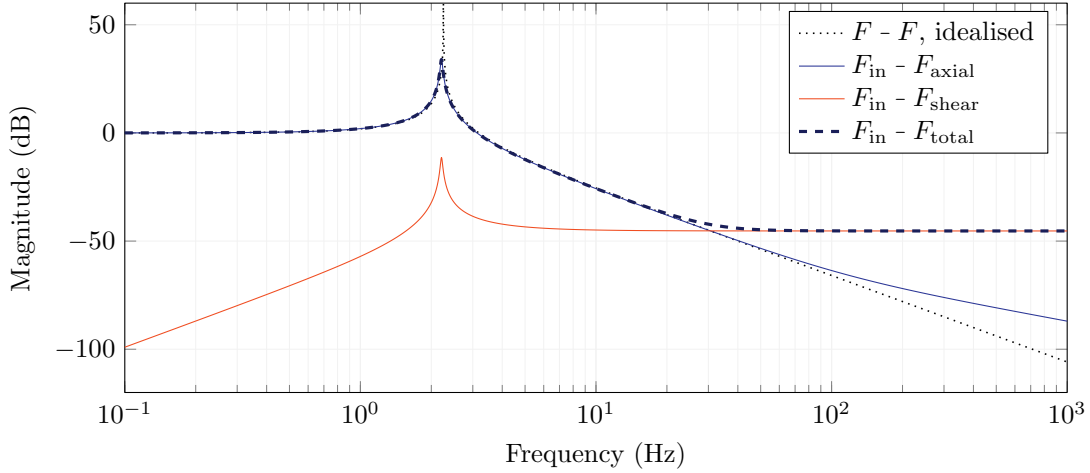

\begin{table}[htb]
\centering
	\setl{tcw}{.1\columnwidth}
	\sisetup{
				table-format=2.1E+4,			
				table-column-width=1.25\tcw, 	
			}
\begin{threeparttable}
\caption{Parameters of the bipod in \Cref{fig: bipod shear}}
\label{tab: 2D pin-pin bipod}	
    \begin{tabular}{L{0.9\tcw}L{1.0\tcw}S[]l}
	\toprule
       								&  Property 	& \multicolumn{1}{l}{Value} 	& Unit 		\\
	\midrule
 	\multirow{9}{*}[-0pt]{Links} 	&	$L$		 	& 0.3 			& \si{\m} 	 				\\
								 	&	${k}$	 	& 5.0E+3		& \si{\N\per\m}				\\
									&	${c}$	   	& 7.2			& \si{\N\s\per\m}		 	\\ [0pt]
									&	$\eta_t$	& 0.7			& 	 						\\
								 	&	$m_t$ 	 	& 0.6			& \si{\kg}					\\
								 	&	$I_t$ 		& 2.5E-3 		& \si{\kg\m\squared}	 	\\ [0pt]
									&	$\eta_b$	& 0.2			& 	 						\\
								 	&	$m_b$	 	& 0.4			& \si{\kg}		 			\\
									&	$I_b$	 	& 1.9E-3		& \si{\kg\m\squared}	 	\\ 
 	\midrule
 	\multirow{1}{*}[-0em]{Payload} 	&	$m_p$		& 25.0			& \si{\kg}					\\
	\bottomrule
    \end{tabular}
\end{threeparttable}
\end{table}

To conclude this section, a comparison of ideal and real struts is shown in \Cref{fig: bipod shear}, with the parameters used summarised in \Cref{tab: 2D pin-pin bipod}. Total force magnitude is extracted from the Lissajous ellipses of the complex vector $\text{TF}_{B_1}+\text{TF}_{B_2}$. The benchmark confirms that link mass and inertia effects could represent an intrinsic drawback severely limiting disturbance attenuation. In fact, although U-joints and S-joints are successfully used for positioning platforms in many practical applications, such BCs are actually counterproductive for microvibration mitigation purposes in which high frequency performance becomes crucial. The next challenge for space missions is to obtain ultrastable satellite platforms while maintaining the spacecraft architecture as simple as possible. For this reason, passive and semi-active hexapod platforms are still considered a key candidate to achieve that, but a thorough analysis on the effect of the boundary conditions is evidently necessary at this point.

\section{Alternative boundary conditions}
\label{sec: alternative BCs}

\subsection{General requirements}

The consideration of different BCs should start from a sufficient condition to obtain an isostatic framework. For a 2D problem, Maxwell's rule in its modern form \cite{calladine:isos_cond} can be expressed as
\begin{equation}\label{eq: maxwell}
 n = r+s-2j
\end{equation}
where $j$, $s$ and $r$ respectively are the number of joints, members and support reactions. For a generic two-dimensional structure,  if $n<0$ the framework is kinematically indeterminate, which means that for an applied load the equilibrium conditions cannot be met at all the joints. Instead, $n \geq 0$ is necessary for determinacy, but is not sufficient, as it does not guarantee mechanism-free behaviour. For kinematically determinate systems, two cases can be distinguished. Firstly, $n=0$ corresponds to a statically determinate structure, that is, member forces and support reactions can be derived directly from the equilibrium. On the other hand, $n>0$ represents a statically indeterminate structure.

\begin{figure}[hbt]
\centering
	\captionsetup[subfigure]{skip=.2\subfigskip}
	\setl{boxw}{0.375\columnwidth}
	\setl{figw}{0.32\columnwidth}
\subcaptionbox{
	\label{fig: diff_bc (a)}}[\boxw]{
	\includegraphics[width=1.0\figw]{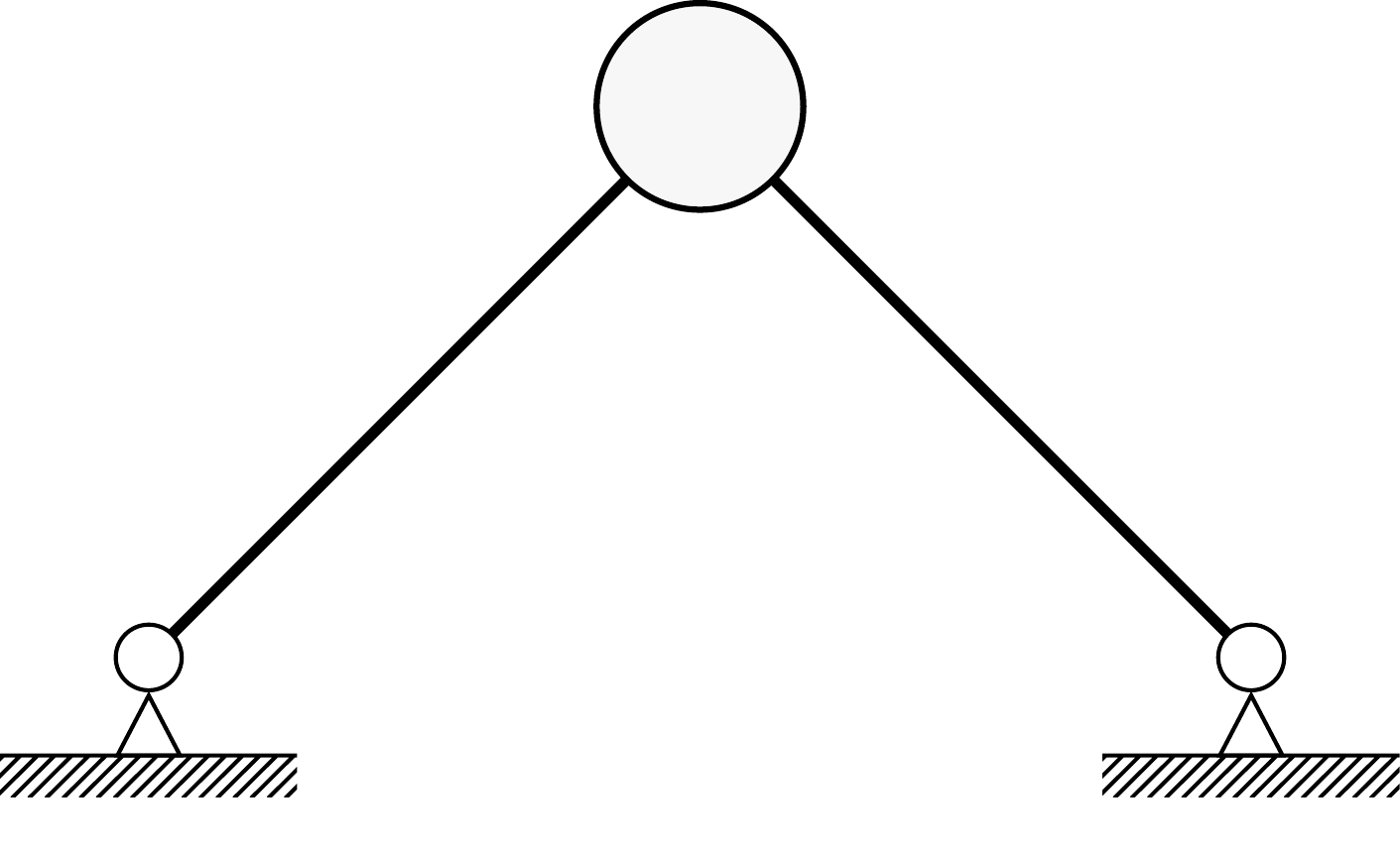}}  					
\subcaptionbox{
	\label{fig: diff_bc (b)}}[\boxw]{
			\includegraphics[width=1.0816\figw]{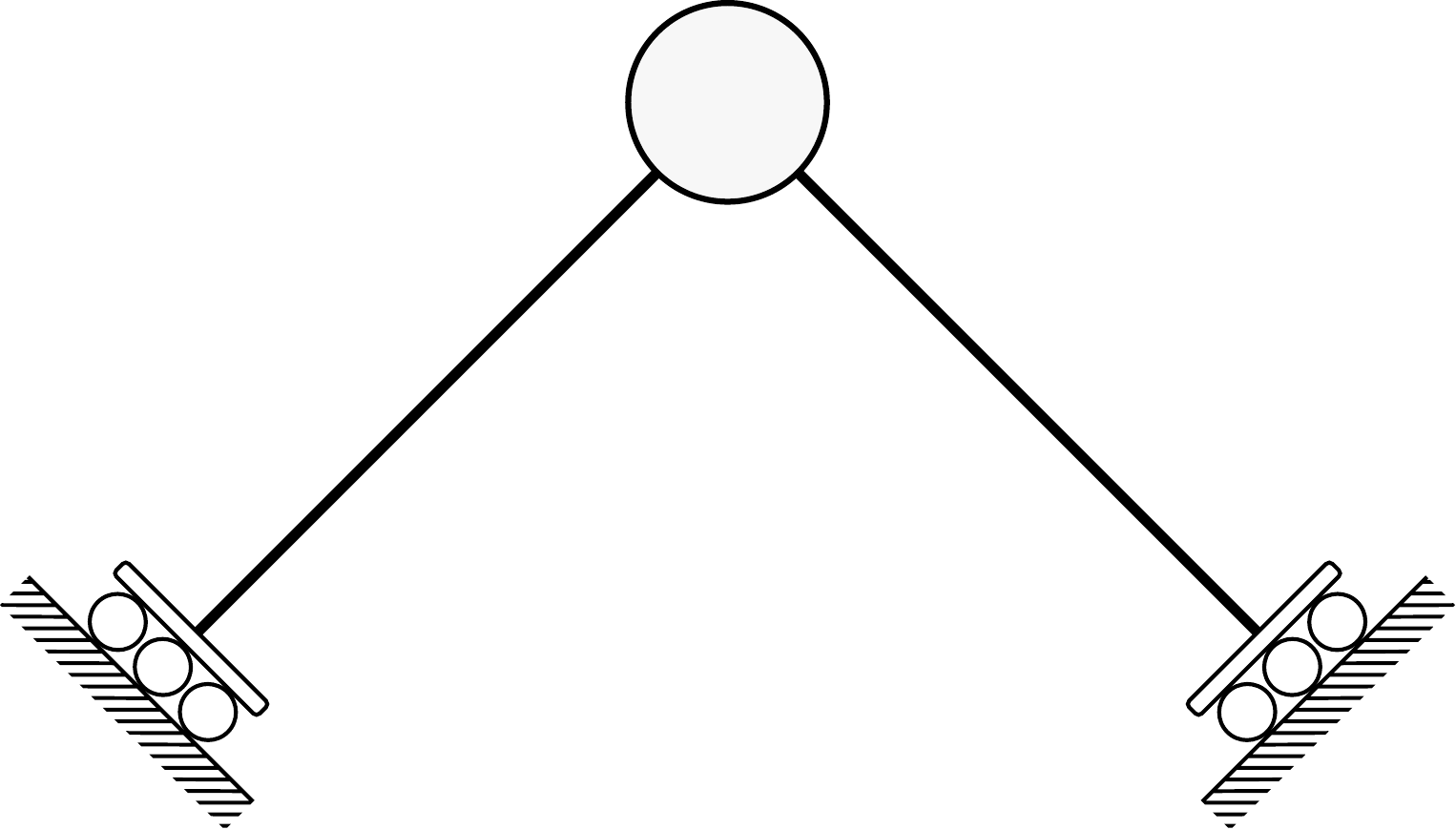}}  			
\\
\vspace{10pt}  															
\subcaptionbox{
	\label{fig: diff_bc (c)}}[\boxw]{
	\includegraphics[width=1.0532\figw]{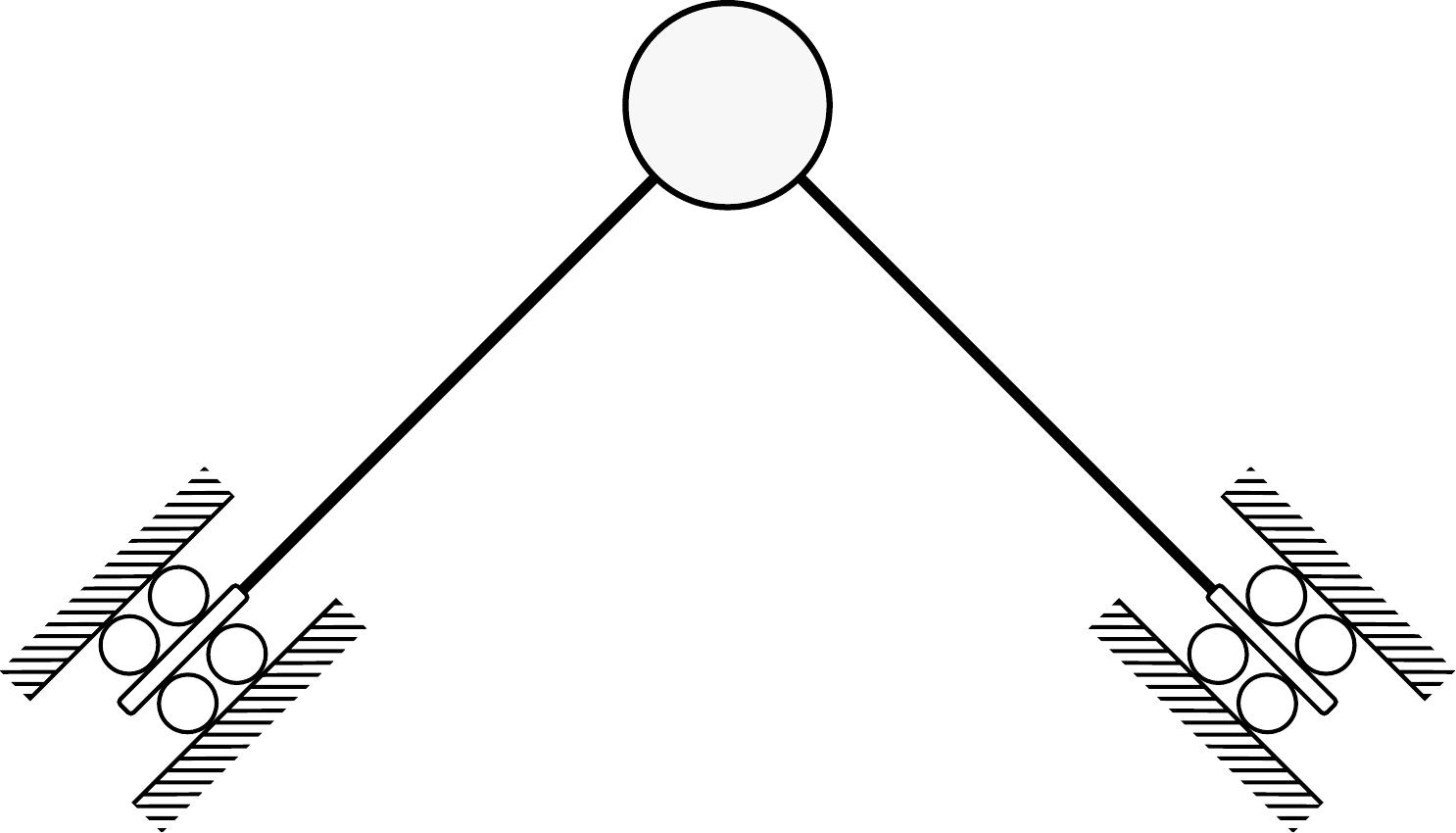}}  					
\subcaptionbox{
	\label{fig: diff_bc (d)}}[\boxw]{
	\begin{tikzpicture}  
	\node[inner sep=0pt, anchor=south west] (img) at (0,0) {\includegraphics[width=1.0532\figw]{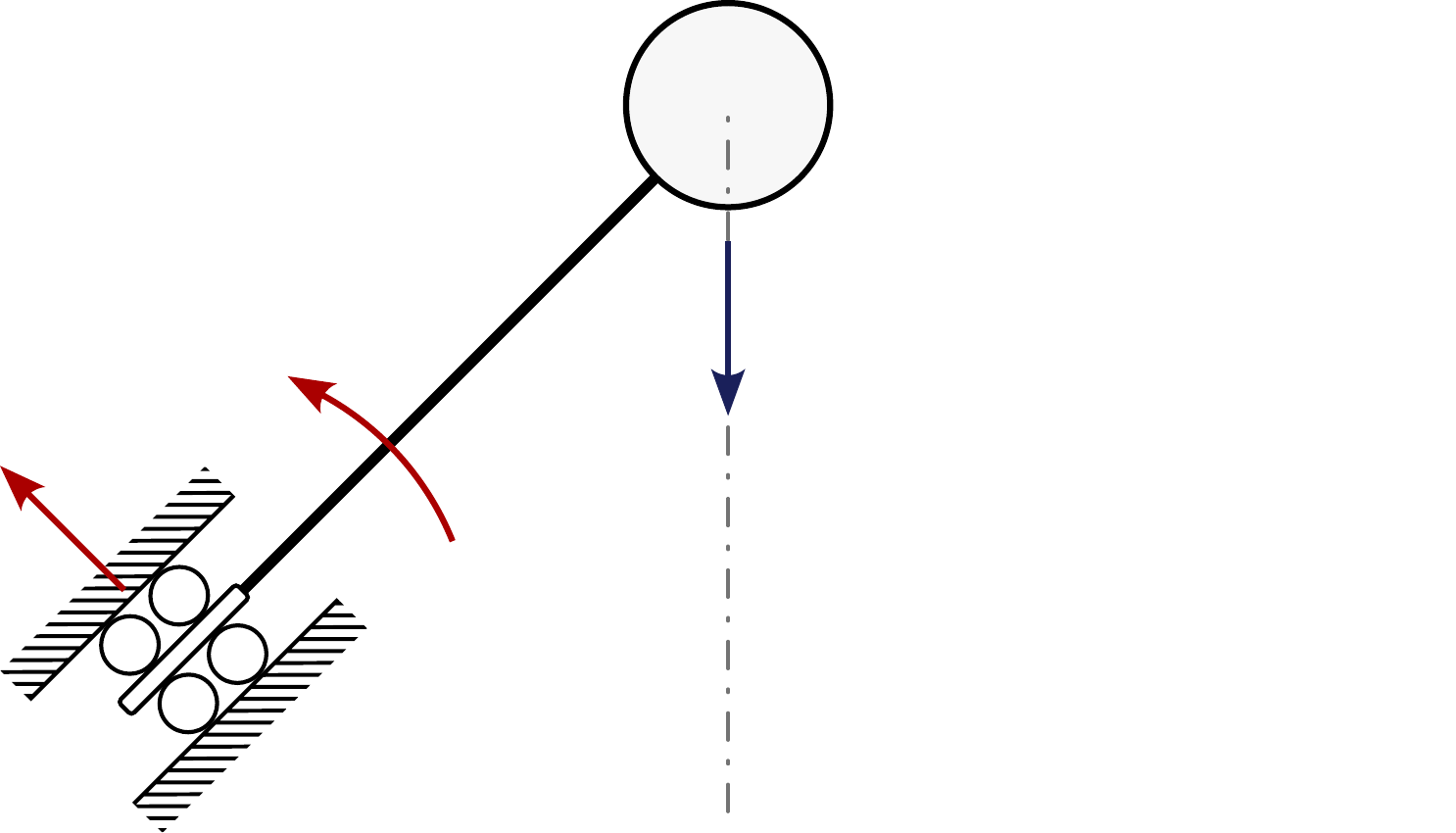}};  
	\begin{scope}[ x={($0.01*(img.south east)$)}, y={($0.01*(img.north west)$)}, ]
		\begin{pgfonlayer}{annotation}
			\node[inner sep=0pt,anchor=south west] at (3.5,41.0) {\small $F_\text{r}$};
			\node[inner sep=0pt,anchor=south west] at (23,54.5) {\small $M_\text{r}$};
			\node[inner sep=0pt,anchor=south west] at (52,56.5) {\small $F_\text{in}$};
		\end{pgfonlayer}
	\end{scope}
	\end{tikzpicture}}	
\caption{Schematic of various symmetric BCs for a bipod: \subref{fig: diff_bc (a)} pin joints, \subref{fig: diff_bc (b)} lateral sliders with \ang{45} inclination to the ground, \subref{fig: diff_bc (c)} longitudinal sliders with \ang{45} inclination to the ground, and \subref{fig: diff_bc (d)} free-body diagram for configuration \subref{fig: diff_bc (c)}}
\label{fig: diff_bc}
\end{figure}
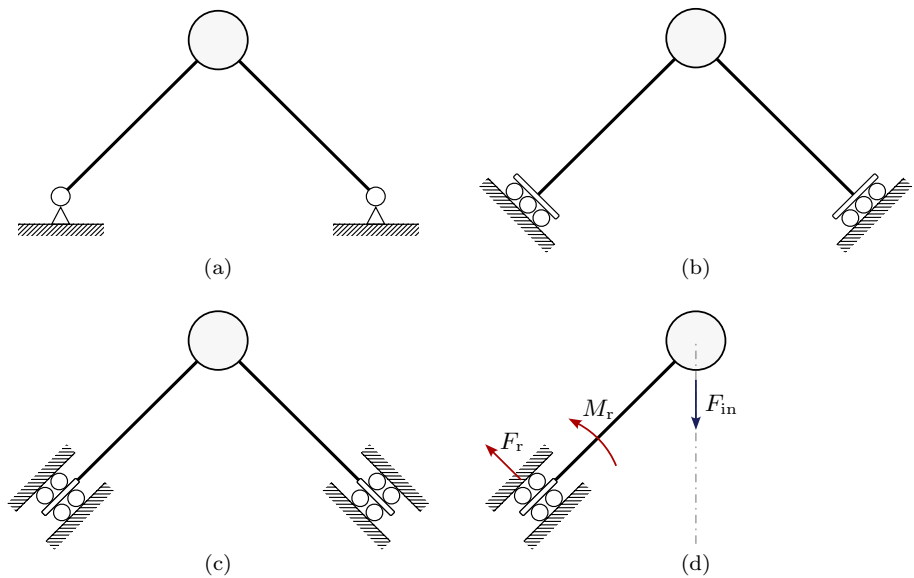

For an uncomplicated symmetric framework, as the one shown in \Cref{fig: diff_bc (a)}, which has 2 members and 3 joints, the condition from \eqref{eq: maxwell} becomes sufficient to assess if the structure is isostatic. In order to have $n=0$, the number of constraint forces needs to be 4, and this is fulfilled by the two pin joints affixed to the ground. To preserve the symmetry and retain the same number of reactions, the two base supports can be substituted with sliders that allow either lateral or longitudinal motion with respect to each strut. \Cref{fig: diff_bc (b)} illustrates the first option, where the sliders are inclined \ang{45} with respect to the ground. The two provided reactions per joint are a moment and an axial force. The latter solution, seen in \Cref{fig: diff_bc (c)}, is characterised by reactions comprising a moment and a shear force. The second option is not viable, because each strut is only loaded transversely, viz. \Cref{fig: diff_bc (d)}. Even if linear springs are introduced in the axial direction, the isostatic condition would be retained without enabling the desired two DOFs. 

Consequently, the pin-slider configuration depicted in \Cref{fig: diff_bc (b)} is chosen as an alternative to the regular pin-pin BC. The roller avoids shear force transmission at the base supports, which was shown in \Cref{ssec: real struts} to hinder system high-frequency isolation capability. It also ensures that struts are exclusively loaded lengthwise. It is worth pointing out that \Cref{fig: diff_bc} serves only for illustrative purposes. Generally, the two bipod links need not be perpendicular, and the roller's motion is taken to always be restricted to the plane orthogonal to its respective link. Formal analysis of the proposed BC is given in \Cref{sec: pin-slider BC}.

\subsection{Pin-slider preliminary case studies}\label{ssec: pin-slider numerical tests}

The viability of the proposed configuration is first studied by two numerical models that were implemented in commercial software. Damping was neglected in these preliminary analyses. Simulink was used for the 2D bipod substructure case, as indicated in \Cref{fig: slider bipod}. The first set of tests conducted aimed at comparing the effect of strut mass on slider-rotational joint systems. In particular, a vertical ($z$-axis) harmonic excitation was applied and the total transmitted force was measured along both the $x$- and $z$-axes. It is seen in \Cref{fig: bipod simulink (a)} that the two cases are almost identical, confirming that link mass does not affect the TF at high frequency when base pins are replaced with lateral sliders. 

A similar conclusion is drawn from \Cref{fig: bipod simulink (b)}, which compares $F_x-F_x$ and $F_x-M_y$ transfer functions for real struts. Here, $F_j$, $M_j$ respectively indicate force and base reaction moment about axis $j$. Force transmission for the all-rotational joint BC exactly matches the predictions from \Cref{ssec: real struts}, whereas the lateral sliders restore the theoretically ideal behaviour. Regarding coupling between input force and output moments, the two boundary conditions perform almost identically below \SI{20}{\Hz}. The pin-pin version ultimately produces about \SI{6}{\dB} lower plateau at higher frequency. Nevertheless, the problem of cross-contamination, that is, the existence of nonzero force-moment coupling TFs, needs to be addressed in a 3D environment. For a parallel manipulator, this effect could be minimised by approaching dynamic isotropy and by strategically placing the centre of mass of the suspended payload.

\begin{figure}[htb]
\centering
	\captionsetup[subfigure]{skip=\subfigskip}
	\setl{boxw}{0.485\columnwidth}
	\setl{figw}{0.485\columnwidth}
\subcaptionbox{
	{Schematic representation}
	\label{fig: slider bipod (a) schema}}[\boxw]{
		\begin{tikzpicture}
		\node[inner sep=0pt, anchor=south west] (img) at (0,0) {\includegraphics[width=1.02\figw]{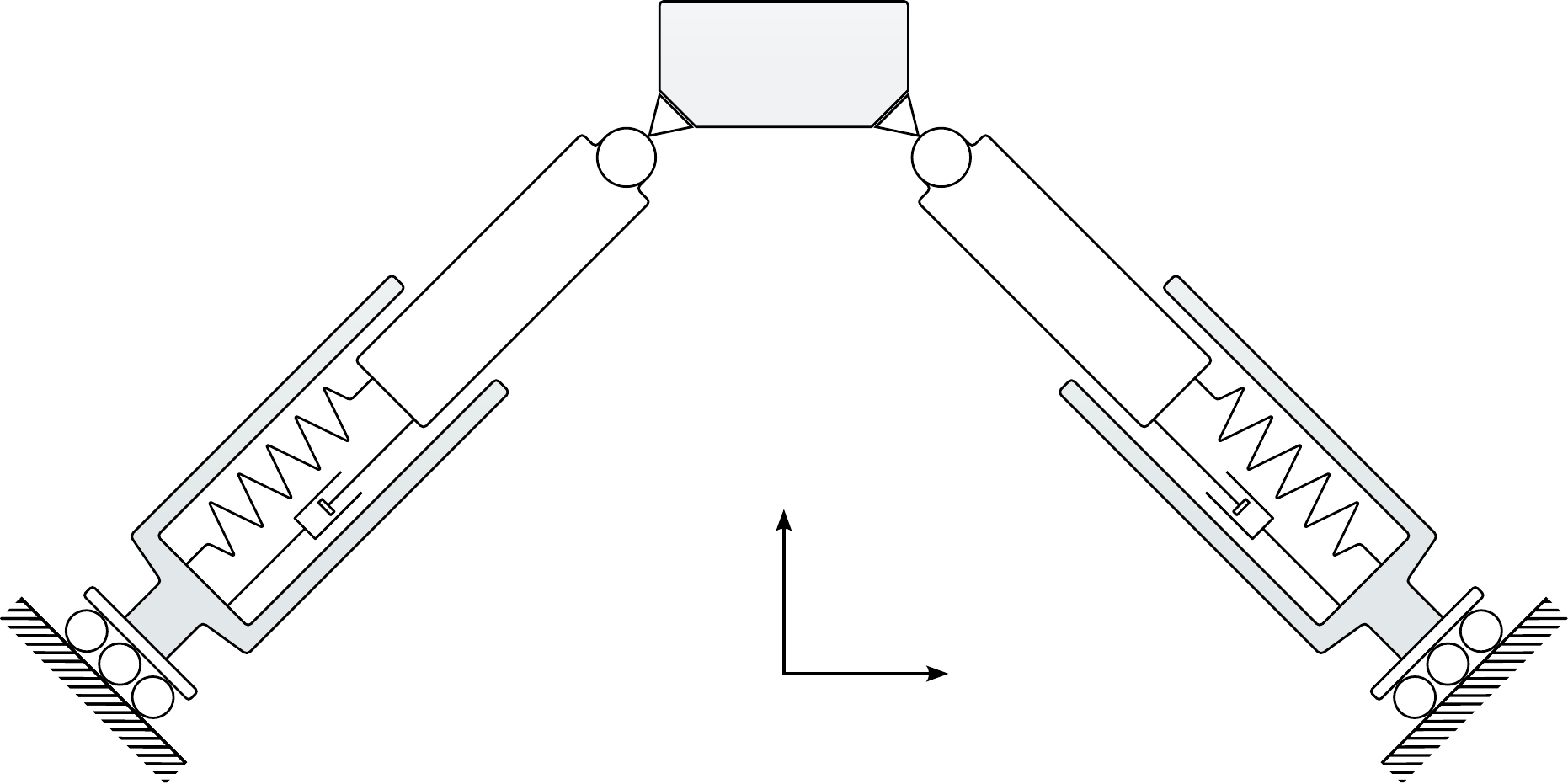}};
		\begin{scope}[ x={($0.01*(img.south east)$)}, y={($0.01*(img.north west)$)}, ]
			\begin{pgfonlayer}{annotation}
				\node[inner sep=1.5pt,anchor=south] at (50.0,35) {$z$};
				\node[inner sep=1.5pt,anchor=west] at (60.5,14.3) {$x$};
				\node[inner sep=-4.5pt,align=center] at (50.0,91.5) {\small $m_p$};
			 	\coordinate (C) at (34.0,63.0);
					\node[inner sep=1pt,anchor=south,rotate around={+45:(C)}] at (C) {\small $m_t$};
			 	\coordinate (D) at (67.0,61.5);
					\node[inner sep=1pt,anchor=south,rotate around={-45:(D)}] at (D) {\small $m_t$};
			 	\coordinate (E) at (16.0,47.0);
					\node[inner sep=3pt,anchor=south,rotate around={+45:(E)}] at (E) {\small $m_b$};
			 	\coordinate (F) at (84.0,47.0);
					\node[inner sep=3pt,anchor=south,rotate around={-45:(F)}] at (F) {\small $m_b$};
			\end{pgfonlayer}
		\end{scope}
	\end{tikzpicture}}
\subcaptionbox{
	{Simulink model}
	\label{fig: slider bipod (b) simulink}}[\boxw]{
	\includegraphics[width=0.95\figw]{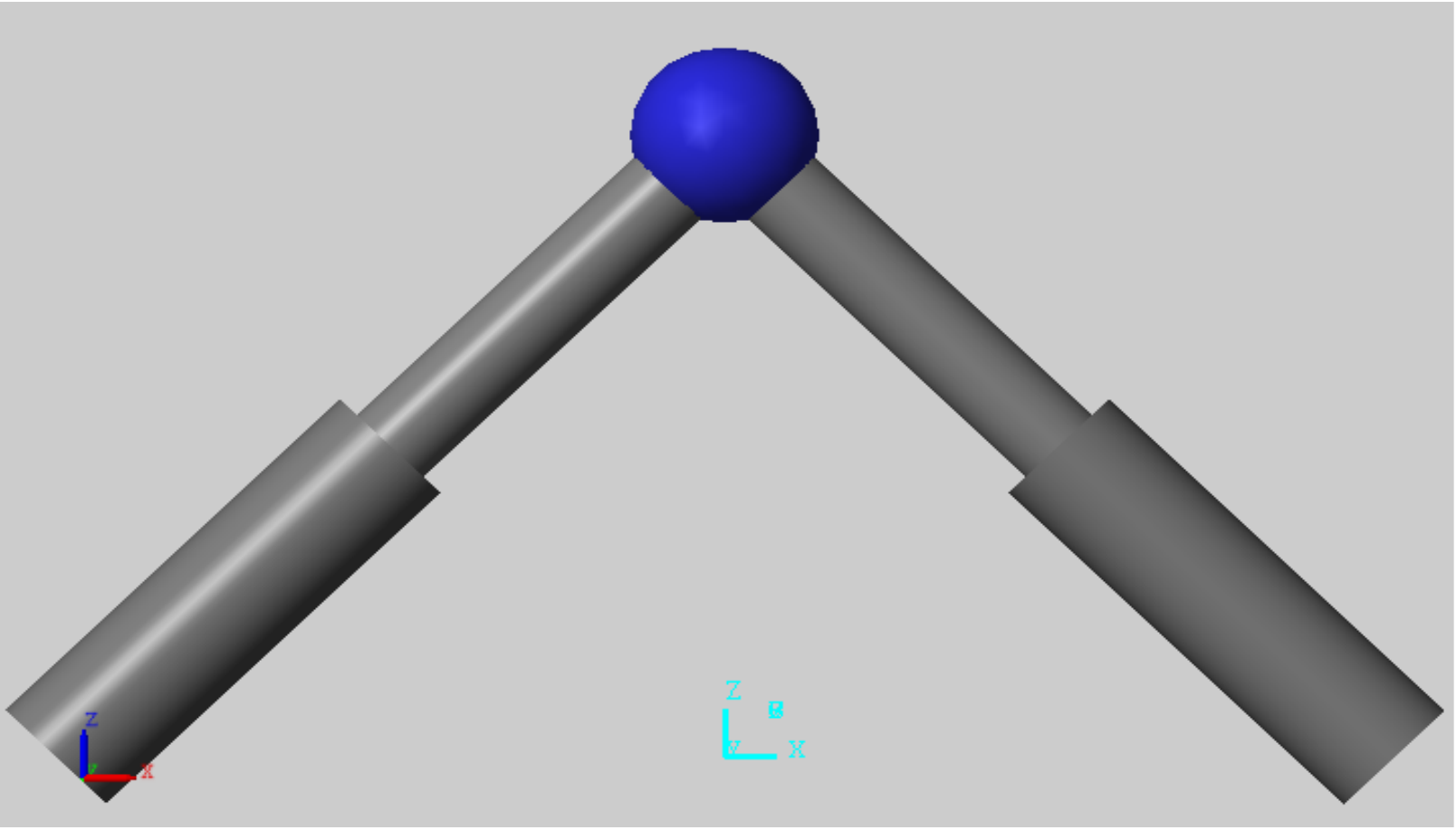}}
\caption{Real struts bipod with BCs of pin at the top and lateral sliders at the base}
\label{fig: slider bipod}
\end{figure}
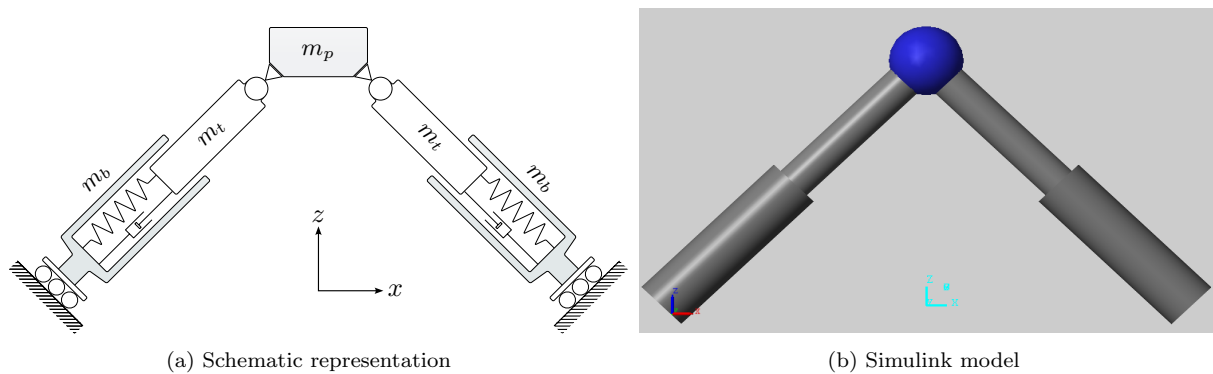

\begin{figure}[htb]
\centering
	\pgfplotsset{every axis/.append style={log subplots, line join=round,}} 
\subcaptionbox{Effect of strut mass on pin-slider BC bipods 
\label{fig: bipod simulink (a)}}[1.0\columnwidth]{
\begin{tikzpicture}
	\begin{axis}[ymin=-200, ymax=60]
	\addplot[rc] table {TikzData/Fz-Fz massless.tsv};
		\addlegendentry{$F_z\shortminus F_z$, massless}
	\addplot[dbc] table {TikzData/Fz-Fz real.tsv};
		\addlegendentry{$F_z\shortminus F_z$, real}
	\addplot[rc,dashed] table {TikzData/Fz-Fx massless.tsv};
		\addlegendentry{$F_z\shortminus F_x$, massless}
	\addplot[dbc,densely dashed] table {TikzData/Fz-Fx real.tsv};
		\addlegendentry{$F_z\shortminus F_x$, real}
	\end{axis}
\end{tikzpicture}
}
\subcaptionbox{Effect of BCs on real strut bipods
\label{fig: bipod simulink (b)}}[1.0\columnwidth]{
\vspace{\plotvsep} 
\begin{tikzpicture}
	\begin{axis}[ymin=-120, ymax=60]
	\addplot[rc] table {TikzData/Fx-Fx all pin.tsv};
	 	\addlegendentry{$F_x\shortminus F_x$, pin-pin}
	\addplot[dbc] table {TikzData/Fx-Fx pin-slider.tsv};
	 	\addlegendentry{$F_x\shortminus F_x$, pin-slider}
	\addplot[rc,densely dashed] table {TikzData/Fx-My all pin.tsv}; 
	 	\addlegendentry{$F_x\shortminus M_y$, pin-pin}
	\addplot[dbc,densely dashed] table {TikzData/Fx-My pin-slider.tsv};
	 	\addlegendentry{$F_x\shortminus M_y$, pin-slider}
	\end{axis}	
\end{tikzpicture}  
}
\caption{Comparison of bipod TFs computed in Simulink between \SI{1}{\Hz} and \SI{1}{\kHz} for different: \subref{fig: bipod simulink (a)} link mass treatment, \subref{fig: bipod simulink (b)} base joint condition. Coordinate system is oriented as in \Cref{fig: slider bipod (a) schema}}
\label{fig: bipod simulink}
\end{figure}

\begin{figure}[H]
\centering
	\captionsetup[subfigure]{skip=\subfigskip}
	\setl{boxw}{0.475\columnwidth} 
	\setl{figw}{0.475\columnwidth} 
\subcaptionbox{
	{Front view}
	\label{fig: adams model (a)}}[\boxw]{
	\includegraphics[width=\figw]{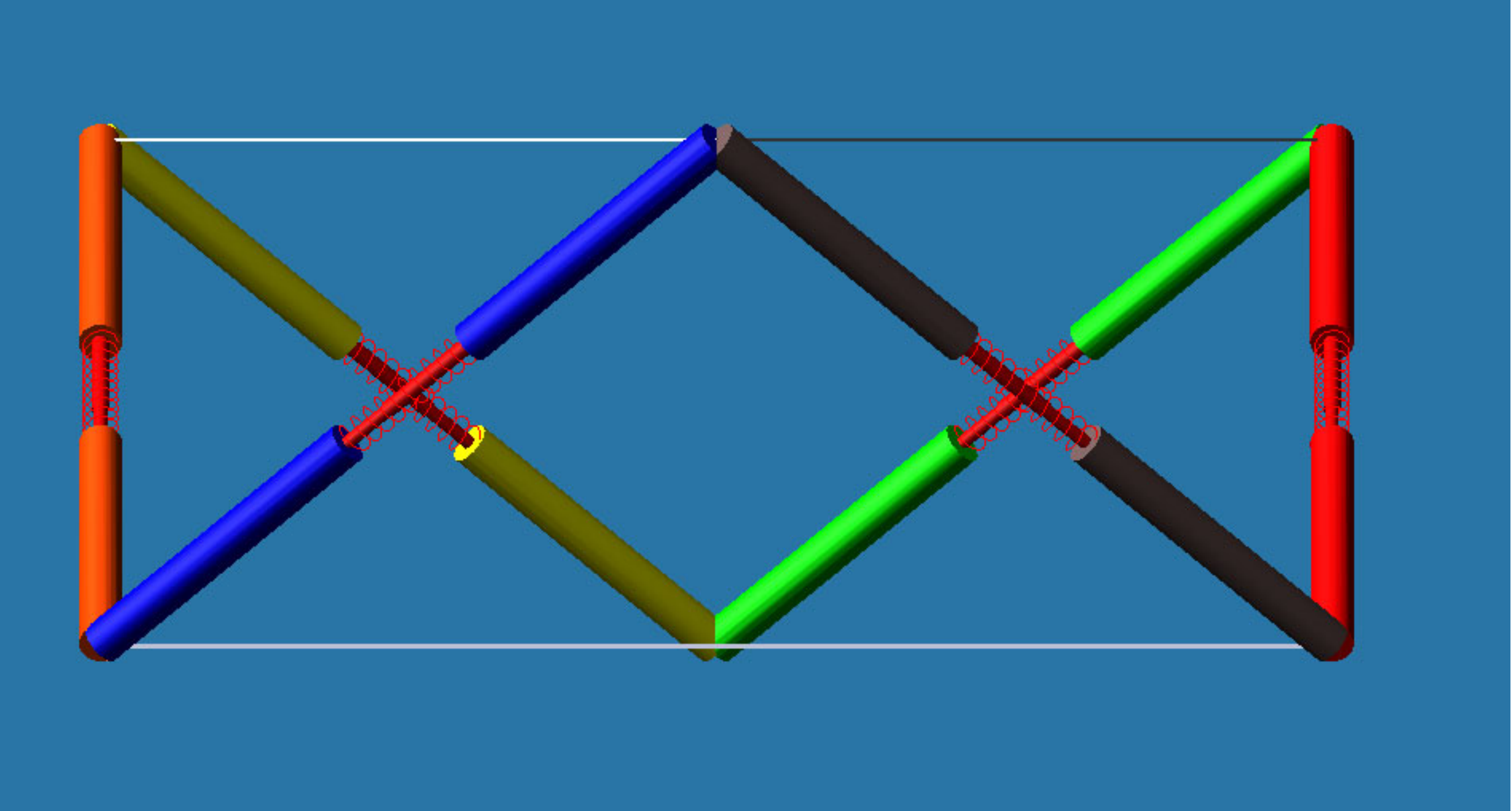}}
\subcaptionbox{
	{Isometric view}
	\label{fig: adams model (b)}}[\boxw]{
	\includegraphics[width=\figw]{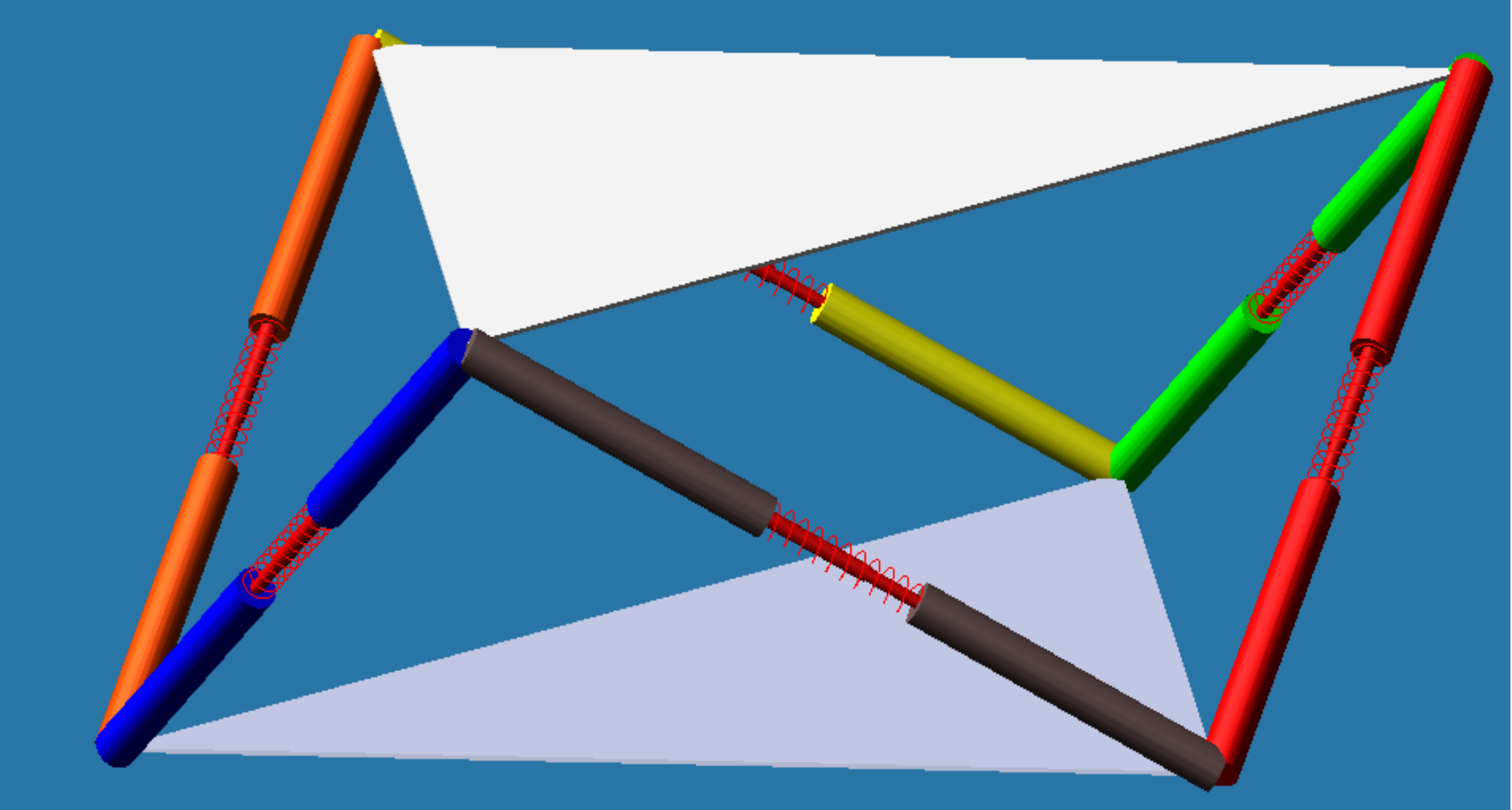}}
\caption{Cubic hexapod model implemented in MSC ADAMS}
\label{fig: adams model}
\end{figure}

As an extension to the preceding planar analysis to 3D, a cubic hexapod model was developed in MCS ADAMS. Analogously, simulations were executed with both types of boundary condition, each employing both ideal and real struts. The model, shown in \Cref{fig: adams model}, uses all rigid bodies. The CM of the suspended element was placed symmetrically between the two platforms. The constraints imposed to obtain six payload DOFs were validated using the mobility formula, also known as  Chebychev–Gr{\"u}bler–Kutzbach criterion. It states that
\begin{equation}\label{eq: mobility formula}
 M = 6n + b - \sum\nolimits_{i=1}^{j}(6-f_i) 
\end{equation}
where $M$ being the number of DOFs, $n$ the moving elements and $b$ the internal DOFs, namely 6 per rigid body. Additionally, $j$ is the number of joints and $f_i$ are the independent DOFs of the $i$th one. Since the base platform is stationary, one has $n=7$, and for a hexapod $j=12$. In order to obtain $M = 6$ from \eqref{eq: mobility formula}, the connections between each strut and the mobile platform were considered to be U-joints, allowing only two rotations. The base attachments each need 3 suppressed DOFs. Therefore, S-joints, which are fully free in rotation and permit no linear motion, complete the all-rotational joint case. For the proposed pin-slider BC, planar joints perpendicular to each link were assigned as the connecting element to the fixed base. They provide 2 in-plane translations and 1 rotation about the strut longitudinal axis. 

Aligned input-output force TFs were collected for pure vertical and lateral excitations, which respectively refer to directions perpendicular and parallel to the platforms' planes at rest. The associated results are plotted in \Cref{fig: adams model (a)} and \subref{fig: adams model (b)}. For either case, the advantages of the pin-roller system over the traditional BC are evident, with both configurations exhibiting very similar dynamic behaviour to the simpler bipod structures investigated previously. In summary, the initial numerical tests suggest that the proposed BC enables the desired high-frequency disturbance attenuation to be achieved. Equations of motion and detailed analysis are presented in the following section.

\begin{figure}[htb]
\centering
	\pgfplotsset{every axis/.append style={log subplots,ymin=-100,ymax=60,line join=round,}} 
\subcaptionbox{Vertical input - vertical output force TF
\label{fig: hexapod adams (a) vert}}[1.0\columnwidth]
{
\begin{tikzpicture}
	\begin{axis}[]
	\addplot[rc,thin] table {TikzData/adams vert spherical massless.tsv}; 	
	 	\addlegendentry{Spherical, massless}
	\addplot[rc,densely dashed] table {TikzData/adams vert spherical real.tsv}; 
	 	\addlegendentry{Spherical, real}
	\addplot[bc,thin] table {TikzData/adams vert planar massless.tsv}; 		 
	 	\addlegendentry{Planar, massless}
	\addplot[dbc,very thick,dashed] table {TikzData/adams vert planar real.tsv}; 
	 	\addlegendentry{Planar, real}
	\end{axis}
\end{tikzpicture}
}
\subcaptionbox{Horizontal input - horizontal output force TF
\label{fig: hexapod adams (b) horiz}}[1.0\columnwidth]
{
\vspace{\plotvsep} 
\begin{tikzpicture}
	\begin{axis}[]
	\addplot[rc,thin] table {TikzData/adams horiz spherical massless.tsv};		 
	 	\addlegendentry{Spherical, massless}
	\addplot[rc,densely dashed] table {TikzData/adams horiz spherical real.tsv};
		 \addlegendentry{Spherical, real}
	\addplot[bc,thin] table {TikzData/adams horiz planar massless.tsv};
	  	\addlegendentry{Planar, massless}
	\addplot[dbc,very thick,dashed] table {TikzData/adams horiz planar real.tsv}; 
	 	\addlegendentry{Planar, real}
	\end{axis}		
\end{tikzpicture}
}
\caption{Effect of strut mass and base platform joint type on transmitted force. Computed in MSC ADAMS from \SI{1}{\Hz} to \SI{1}{\kHz} for a cubic hexapod. TFs displayed are for aligned input-output: \subref{fig: hexapod adams (a) vert} vertical, \subref{fig: hexapod adams (b) horiz} horizontal}
\label{fig: hexapod adams}
\end{figure}
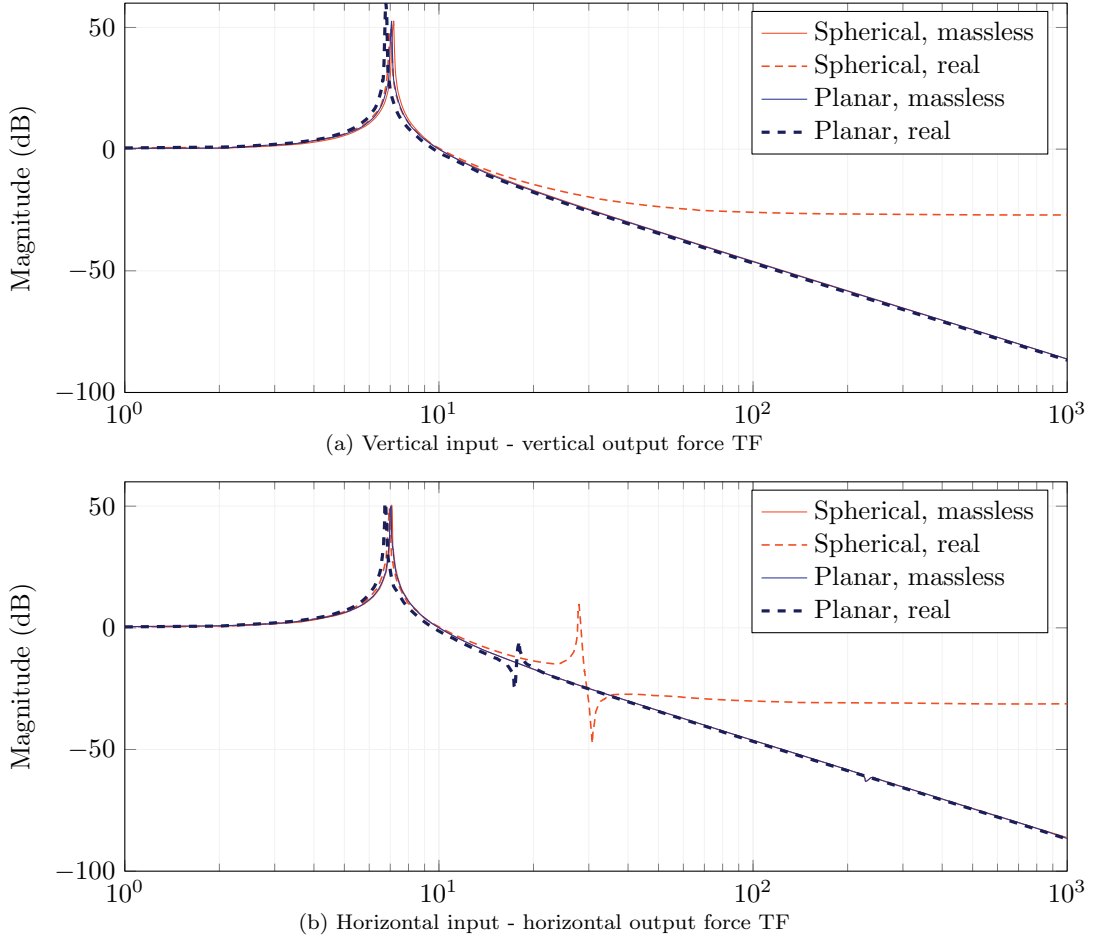

\section{Pin-slider BC analysis}
\label{sec: pin-slider BC}

\subsection{Equations of motion}\label{ssec: equations of motion}

In this section, Newton-Euler equations of motion for pin-slider parallel manipulators are derived. Initially, it is assumed that the system is composed of any number of perpendicular strut bipods. This case can be conveniently described with fewer coordinate systems. A trivial modification that enables treatment of parallel manipulators made up of arbitrarily arranged links is supplied in \Cref{ssec: extension to arbitrary parallel manipulators}. 

\begin{figure}[h]
 \centering
 \includegraphics[width=.8\columnwidth]{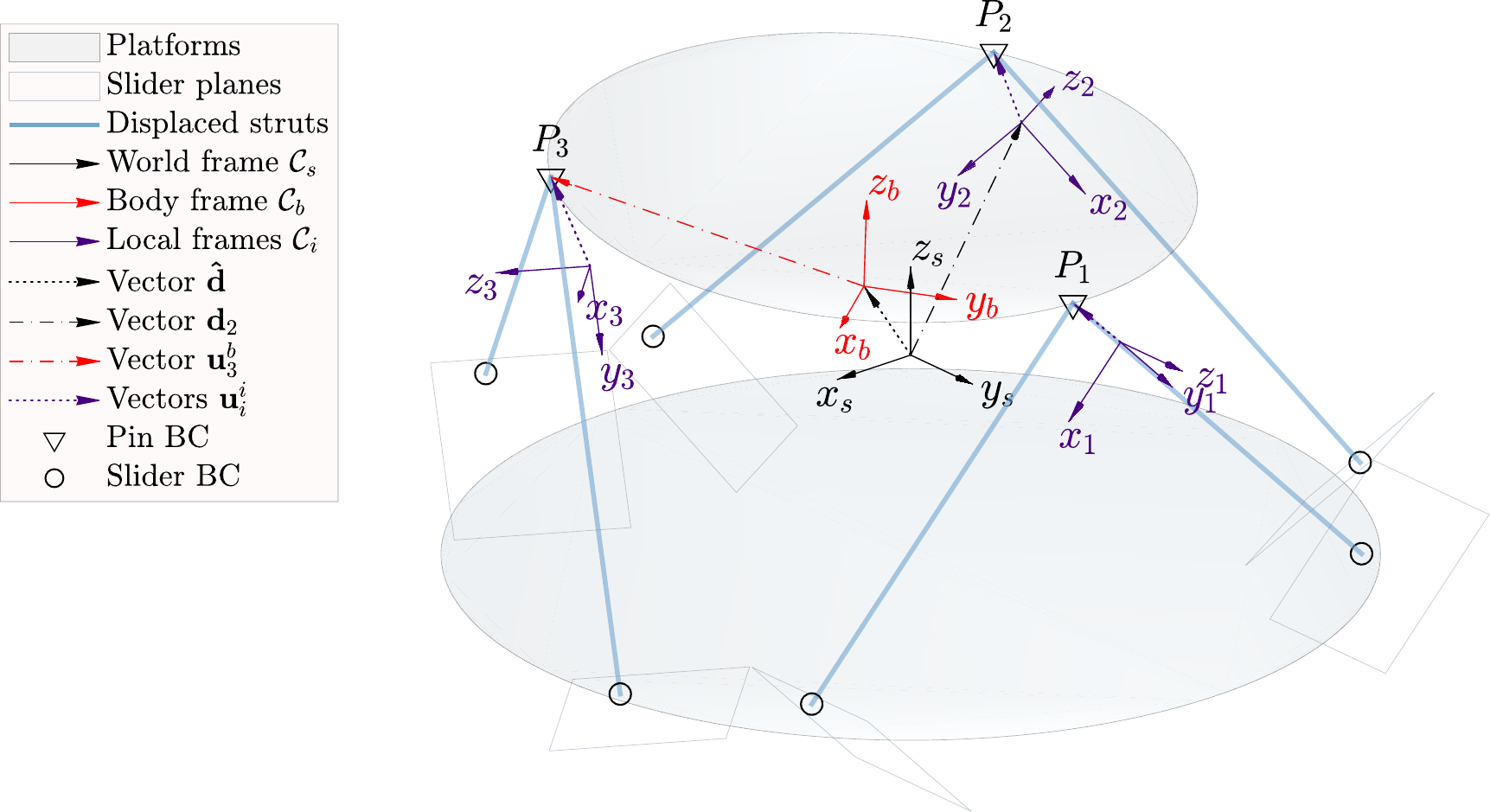}
 \caption{Coordinate systems of a displaced pin-slider hexapod. Nominal slider positions are the centres of the rectangles marking the respective constraint planes. In this example payload principal axes are not aligned with $\csys{C}_s$ at rest}
 \label{fig: hexapod displaced}
\end{figure}

Let $\csys{C}_s$ be an inertial frame with origin the nominal position of the equipment's CM. Orientation is defined so that axis $y_s$ is parallel to the plane of symmetry one bipod and axis $z_s$ perpendicular to the hexapod base and directed away from it. Additionally, the body-fixed $\csys{C}_b$ is attached to the payload and aligned with its principal axes of rotation. Abiding to an anticlockwise strut numbering, a local inertial frame $\csys{C}_{i}$ is constructed for the ${i}$th bipod. Their origins are the bipod pin joint points $P_{i}$ at rest, whose space frame coordinates are $\vc{d}_{i}$. Axes $x_{i}$, $y_{i}$ are strut-aligned as depicted in \Cref{fig: hexapod displaced}, whereas $z_{i}$ is directed away from the upper platform. Let $\map{T}_{j\rightarrow q}:\mathbb{R}^3 \to \mathbb{R}^3$ denote an affine transformation from $\csys{C}_j$ to $\csys{C}_q$, composed of rotation and translation. Set
\begin{subequations}\label{eq: T_bs,T_ks}
 \begin{align}
  \map{T}_{b\rightarrow s}(\vc{x}) 		&= \m{\hat{R}}\vc{x} + \vc{\hat{d}}  						  	\label{eq: T_bs} \\ 
  \map{T}_{{i}\rightarrow s}(\vc{x}) 	&= \m{R}_{i}\vc{x} + \vc{d}_{i}  	,\quad {i} = 1,2,\dots,n    \label{eq: T_ks}
 \end{align}
\end{subequations}
The time dependence of $\map{T}_{b\rightarrow s}$, $\m{\hat{R}}$ and $\vc{\hat{d}}$ is omitted to simplify notation, while  $\m{R}_{i}$ and $\vc{d}_{i}$ are constants. Superscript $j$, for some $j\neq s$, is introduced to indicate a quantity represented in $\csys{C}_j$. No superscript implies the world frame $\csys{C}_s$. Finally, subscript $0$ denotes the isolation system reference configuration.

\noindent
The corotational and bipod coordinates of joint $P_{i}$ are given by
\begin{subequations}\label{eq: ukb, ukk}
 \begin{align}
  \vc{u}_{i}^b 	&= \inv{\map{T}}_{b\rightarrow s \text{,\,}0}(\vc{d}_{i}) = \mt{\hat{R}}_0\vc{d}_{i}	\label{eq: ukb ukk (a)} \\
  \vc{u}_{i}^{i}&= \map{T}_{s\rightarrow {i}} ( \map{T}_{b\rightarrow s} ( \vc{u}_{i}^b ) )  
	   		 	 = \inv{\map{T}}_{{i}\rightarrow s}  (\vc{u}_{i}) 
 			 	 = \mt{R}_{i} ( \vc{u}_{i} - \vc{d}_{i} ) 											 	\label{eq: ukb ukk (b)}
 \end{align}
\end{subequations}
where $\vc{u}_{i}^b$ is constant and $\vc{u}_{i}^{i}$ is the displacement of $P_{i}$ in $\csys{C}_{i}$. When equipment principal axes at rest coincide with $\csys{C}_s$, one has $\m{\hat{R}}_0 = \m{E}$, with $\m{E}$ being the identity matrix. Let $\skewt{.}$ be the skew-symmetric operator,
\begin{equation}\label{eq: [.]x operator}
 \skewt{\vc{w}} = 
 \begin{bmatrix}
  0  		& -w_3   	&  w_2  	 \\
  w_3 		& 0  	  	&  -w_1 	 \\
  -w_2  	& w_1  		&  0 
 \end{bmatrix} 
 ,\quad  \vc{w} = \begin{bmatrix} w_1 \\ w_2 \\ w_3 \end{bmatrix}
\end{equation}
It converts cross products to matrix-vector multiplication as $\vc{w}\times\vc{z} = \skewt{\vc{w}}\vc{z}$ for any $\vc{w},\vc{x}\in\mathbb{R}^3$. Denoting angular velocity by $\vc{\omega}^b$ and differentiating \eqref{eq: ukb ukk (b)} gives the motion of $P_{i}$ in $\csys{C}_{i}$
\begin{subequations}\label{eq: ukk}
 \begin{align}
  \vc{u}_{i}^{i} 		 	
   &= \mt{R}_{i} \big( \vc{\hat{d}} + \m{\hat{R}}\vc{u}_{i}^b - \vc{d}_{i} \big) 	
\label{eqs: ukk}	\\
  \dot{\vc{u}}_{i}^{i} 	
   &= \mt{R}_{i} \big( \dot{\hat{\vc{d}}} + \m{\hat{R}}\skewt{\vc{\omega}^b}\vc{u}_{i}^b  \big)	
\label{eq: ukk dot}	\\					
  \ddot{\vc{u}}_{i}^{i} 	
   &= \mt{R}_{i} \big( \ddot{\hat{\vc{d}}} + \m{\hat{R}}\skewt{\dot{\vc{\omega}}^b}\vc{u}_{i}^b 
      + \m{\hat{R}} \skewt{\vc{\omega}^b}^2\vc{u}_{i}^b  \big)
\label{eqs: ukk ddot}
 \end{align}
\end{subequations}

The kinematics of bipod substructures are examined first. The planar joint BC permits each roller to move within a plane orthogonal to its link's axis, see \Cref{fig: hexapod displaced}. The sliders have two translational DOFs, namely $x_{i}$, $z_{i}$ for the $y_{i}$-aligned strut and $y_{i}$, $z_{i}$ for the $x_{i}$ one. Save for each link's respective longitudinal axis, this BC fully eliminates base rotations, whereas the top pin joint does not admit input moments. The angle between the two struts is therefore fixed and the out-of-plane motion of the bipod, i.e. along $z_{i}$, is confined to uniform translation of the whole substructure. The assertion implies that strut masses undergo no rotation in inertial frames. Following the strut property annotation from \Cref{ssec: real struts}, $P_{i}$ joint force is 
\begin{equation}\label{eq: 2D governing eq}
 \vc{f}_{P,{i}}^{i} = - {k}\m{P}\vc{u}_{i}^{i} - {c}\m{P}\dot{\vc{u}}_{i}^{i} - m_s\m{S}\ddot{\vc{u}}_{i}^{i}
\end{equation}	
where the orthogonal projector $\m{P}$ and mass scaling $\m{S}$ are
\begin{equation}\label{eq: 2D governing eq P, S}
 	\m{P} =  \diag(1 ,\: 1 ,\:  0)  				,\quad 
	\m{S} =  2\m{E} - \frac{m_b}{m_s} \m{P}
\end{equation}	
and $\diag$ is the vector-to-matrix operator. The full parallel manipulator dynamics are characterised next. Using \eqref{eq: 2D governing eq} and summing the contributions of all joints, the total $\csys{C}_s$ force and $\csys{C}_b$ torque about the equipment's centre of mass become
\begin{subequations}\label{eq: reqction F,tau in Cb}
 \begin{align}
  \vc{f}_{P,\text{tot}} 	 	& = \sum_{i=1}^{n}\nolimits \m{R}_{i} \vc{f}_{P,{i}}^{i} 						\\
  \vc{\uptau}_{P,\text{tot}}^b 	& = \sum_{i=1}^{n}\nolimits \skewt{\vc{u}_{i}^b} \mt{\hat{R}} \m{R}_{i} \vc{f}_{P,{i}}^{i}
 \end{align}
\end{subequations}
The platform experiences disturbances $\vc{f}_\text{in}^b$, $\vc{\uptau}_\text{in}^b$ stemming from payload operation, along with the reactions established in \eqref{eq: reqction F,tau in Cb}. Applying the linear and angular momentum conservation laws about the CM in $\csys{C}_s$ and $\csys{C}_b$, respectively, gives the equations of motion
\begin{subequations}\label{eq: conserv. laws}
 \begin{align}
  \m{\hat{R}}\vc{f}_\text{in}^b + \vc{f}_{P,\text{tot}}	 &= m_p\ddot{\hat{\vc{d}}}                                                       
 	\label{eq: conserv. laws F} \\ 
  \vc{\uptau}_\text{in}^b + \vc{\uptau}^b_{P,\text{tot}} &= \m{I}^b \dot{\vc{\omega}}^b  + \skewt{\vc{\omega}^b} \m{I}^b \vc{\omega}^b  
 	\label{eq: conserv. laws M}
 \end{align}
\end{subequations}
The mass $m_p$ and inertia tensor $\m{I}^b$ refer only to the equipment suspended on the bipods. Labelling principal moments of inertia $I_j$ the latter reduces to $\m{I}^b = \diag(I_x,\: I_y,\: I_z)$. Any rotations convention may be used to specify $\m{\hat{R}}$ and $\vc{\omega}^b$. For example, adopting the 3-2-1 Tait-Bryan angle set gives
\begin{equation}\label{eq: 3-2-1 Rhat}
\m{\hat{R}}(\psi,\theta,\phi) = 
 \begin{bmatrix}
  \cA 	& -\sA 	 & 0  	\\
  \sA 	& \cA	 & 0  	\\
  0   	& 0 	 & 1
 \end{bmatrix}
 \begin{bmatrix}
  \cB  	& 0 	& \sB  	\\
   0   	& 1 	& 0  	\\
  -\sB 	& 0 	& \cB
 \end{bmatrix} 
 \begin{bmatrix}
  1 	& 0 	& 0  	\\
  0 	& \cC 	& -\sC  \\
  0   	& \sC	& \cC
 \end{bmatrix}  
\end{equation}
Angular velocity is related to $\psi$, $\theta$, $\phi$ and their derivatives via
\begin{equation}\label{eq: Euler basis}
 \m{B} = 
 \begin{bmatrix}
  -\sB 		& 0 	& 1	  \\
  \cB\sC 	& \cC 	& 0   \\
  \cB\cC  	& -\sC	& 0
 \end{bmatrix} 
,\quad
 \vc{\alpha} = \begin{bmatrix} \psi \\ \theta \\ \phi \end{bmatrix}		 
,\quad	
  \vc{\omega}^b = \m{B} \dot{\vc{\alpha}}	
\end{equation}

Introducing a state vector $\vc{y}$ allows the Newton-Euler equations \eqref{eq: conserv. laws} to be reformulated as a first order nonlinear \gls{acr:ode}   
\begin{subequations}\label{eq: state space def}
 \begin{align}
 \vc{y} &= \big[\begin{matrix} \vc{\hat{d}} & \vc{\alpha} & \dot{\hat{\vc{d}}} & \vc{\omega}^b	\end{matrix}\tr{\big]}		\\
 \dot{\vc{y}} &= \mpi{M}\vc{g}   
 \end{align}
\end{subequations}
where superscript $+$ indicates the pseudoinverse and $\m{M}$, $\vc{g}$ are functions of $t$, $\vc{y}$. Suitable algebraic manipulations of \eqref{eq: ukk}-\eqref{eq: Euler basis}, elaborated in \ref{sec: apx 3D deriv}, lead to 
\begin{equation}\label{eq: state space M,y,g}
\m{M} = 
 \begin{bmatrix}
	\m{E} & & &  \\
	& \m{B}  & &  \\
	& & m_p\m{E} + m_s \sum_{i}\nolimits \m{\bar{S}}_{i}  & -m_s \m{\bar{D}} \\
	&  & -m_s \mt{\bar{D}} & -m_s\m{\bar{U}}^b + \m{I}^b
 \end{bmatrix}
,\quad
\vc{g} = 
 \begin{bmatrix}
	\dot{\hat{\vc{d}}}  														\\
	\vc{\omega}^b																\\
	\m{\hat{R}}\vc{f}_\text{in}^b  -  \sum_{i}\nolimits \vc{\bar{f}}_{P,{i}}			\\
	\vc{\uptau}_\text{in}^b  - \sum_{i}\nolimits \vc{\bar{\uptau}}_{P,{i}}^b  
	- \skewt{\vc{\omega}^b} \m{I}^b \vc{\omega}^b
 \end{bmatrix}
\end{equation}
with
\begin{equation}\label{eq: state space vars}
\begin{aligned}[c]
 	\m{\bar{S}}_{i} &= \m{R}_{i}\m{S}\mt{R}_{i} 																			\\						
 	\m{\bar{D}} &= \sum_{i=1}^{n}\nolimits \m{\bar{S}}_{i}\m{\hat{R}}\skewt{\vc{u}_{i}^b} 										\\
 	\m{\bar{U}}^b &= \sum_{i=1}^{n}\nolimits \skewt{\vc{u}_{i}^b} \mt{\hat{R}}\m{\bar{S}}_{i}\m{\hat{R}}  \skewt{\vc{u}_{i}^b}	\\
 	\vc{\bar{f}}_{P,{i}} &= \m{R}_{i}\m{P}  \big( {k}\vc{u}_{i}^{i}  +  {c}\dot{\vc{u}}_{i}^{i}  \big)
		+ m_s \m{\bar{S}}_{i} \m{\hat{R}} \skewt{\vc{\omega}^b}^2\vc{u}_{i}^b												\\
 	\vc{\bar{\uptau}}_{P,{i}}^b &= \skewt{\vc{u}_{i}^b} \mt{\hat{R}} \vc{\bar{f}}_{P,{i}}
\end{aligned}
\end{equation}
where the expressions for $\vc{u}_{i}^b$, $\vc{u}_{i}^{i}$ and $\dot{\vc{u}}_{i}^{i}$ are respectively taken from \eqref{eq: ukb ukk (a)} and \eqref{eq: ukk}. It should be mentioned that $\m{M}(t,\vc{y})$ is nonsingular for physically meaningful systems, thus $\mpi{M}$ coincides with $\mi{M}$.

\subsection{Extension to arbitrary parallel manipulators}
\label{ssec: extension to arbitrary parallel manipulators}

Foregoing the assumption of orthogonal strut bipods, \eqref{eq: 2D governing eq} can be used to describe only a single link. Let $\vc{e}_{ j }^p$ denote the unit vector of axis ${ j }$ in $\csys{C}_p$. It is sufficient to construct a separate coordinate system $\csys{C}_{i}$ aligned by $\vc{e}_{ j }^{i}$ for each link and redefine $\m{P}$ and $\m{S}$ as
\begin{equation}\label{eq: 2D governing eq P, S V2}
	\m{P} = \diag(\vc{e}_{ j }^{i}) ,\quad 
	\m{S} = \m{E} - \frac{m_b}{m_s} \m{P}
\end{equation}
e.g. for $x_{i}$-aligned struts, $\vc{e}_{x_{i}}^{i} = [\begin{matrix} 1 & 0 & 0 \end{matrix}]\tr{}$, $\m{P}=\diag(1 ,\: 0 ,\: 0)$ and $\m{S}=\diag(m_t ,\: 1 ,\: 1)$. In other words, the analysis in \Cref{sec: pin-slider BC} remains valid when substituting \eqref{eq: 2D governing eq P, S V2} in place of \eqref{eq: 2D governing eq P, S} and having $n$ as the number of links of the parallel manipulator as opposed to the number of bipods. 

The benefit of the bipod-based approach is that whenever it is applicable, such as for the hexapod in \Cref{fig: hexapod displaced}, the system dynamics can be characterised with half the number of $\csys{C}_{i}$ frames and less superfluous computation. All preceding derivations are generic, in the sense that they are not bound to any particular system geometry. For example, with $n=1$, $\vc{d}_1 = \vc{0}$ and using \eqref{eq: 2D governing eq P, S}, the equations of motion automatically collapse to a pin-slider version of the bipod from \Cref{ssec: real struts}. An equivalent representation is obtained with $n=2$, $\vc{d}_1 = \vc{d}_2 = \vc{0}$ and \eqref{eq: 2D governing eq P, S V2}. In addition, the system need not be rotationally symmetric, i.e. using the more general formulation \eqref{eq: 2D governing eq P, S V2} with $n=1$, the ODE in \eqref{eq: state space def} pertains to a payload attached to a single strut. For the special case of ideal struts, all $m_s$-dependent terms trivially vanish.

\subsection{Transmitted forces and moments}
\label{ssec: transmitted forces and moments}
At this stage, the individual contributions $\vc{f}_{\text{out},{i}}$ and total force $\vc{f}_\text{out}$ transmitted to base can be extracted. A straightforward summation results in
\begin{equation}\label{eq: f_out}
 \vc{f}_\text{out}  = \sum_{i=1}^{n}\nolimits \vc{f}_{\text{out},{i}} 
 					= \sum_{i=1}^{n}\nolimits \m{R}_{i} \m{P} \big( {k}\vc{u}_{i}^{i} + {c}\dot{\vc{u}}_{i}^{i} \big) 	
\end{equation}
Evaluation of $\vc{u}_{i}^{i}$ and $\dot{\vc{u}}_{i}^{i}$ is done by substituting computed state vector values into \eqref{eq: ukk}. Ground constraint moment $\vc{\upmu}_{\text{r},G}$ for the whole system about a reference point $G$ is obtained as follows. Let $\vc{g}$ and $\vc{s}_{i}$ denote the space frame coordinates of $G$ and the ${i}$th slider, respectively. The latter's relative displacement in $\csys{C}_i$ is $(\m{E}-\m{P}) \vc{u}_{i}^{i}$, whereas its position at rest is $L\vc{e}_{ j }^{i}$. Then
\begin{subequations}
	\begin{align}
		\vc{\upmu}_{\text{r},G} 	&= -\sum_{i=1}^{n}\nolimits  \skewt{\vc{g}-\vc{s}_{i}}  \vc{f}_{\text{out},{i}} 				\\
		\vc{s}_{i}		 			&= \m{R}_{i}\big[ (\m{E}-\m{P}) \vc{u}_{i}^{i}  + L\vc{e}_{ j }^{i} \big]  + \vc{d}_{i}
	\end{align}
\end{subequations}

Reaction moments $\vc{\upmu}_{\text{r},{i}}$ of each planar joint can also be retrieved. Analogously to \Cref{ssec: real struts}, the normal distance from a roller's plane of motion to its respective link's CM is specified by a nondimensional constant $\eta_s = (\eta_t m_t + \eta_b m_b)/m_s$. In local coordinates 
\begin{equation}\label{eq: tau_r}
 \vc{\upmu}_{\text{r},{i}}^{i} 	= \skewt{\eta_s L\vc{e}_{ j }^{i}} m_s \m{S}\ddot{\vc{u}}_{i}^{i}
							 	 	= \eta_s L m_s  \skewt{\vc{e}_{ j }^{i}} \ddot{\vc{u}}_{i}^{i}
\end{equation}
noticing that $\skewt{\vc{e}_{ j }^{i}}\m{S} = \skewt{\vc{e}_{ j }^{i}}$ follows from \eqref{eq: 2D governing eq P, S V2}.

\subsection{Hexapod geometry parametrisation}
\label{ssec: hexapod geometry parametrisation}

\begin{figure}[htb]
	\centering
	\captionsetup[subfigure]{skip=1.0\subfigskip}
	\setl{boxw}{0.4\columnwidth}
	\setl{figw}{0.375\columnwidth}
	\subcaptionbox{
		{Top view}
		\label{fig: geom parametrisaiton (a)}}[\boxw]{
		\begin{tikzpicture}
			\node[inner sep=0pt, anchor=south west] (img) at (0,0) {\includegraphics[width=\figw]{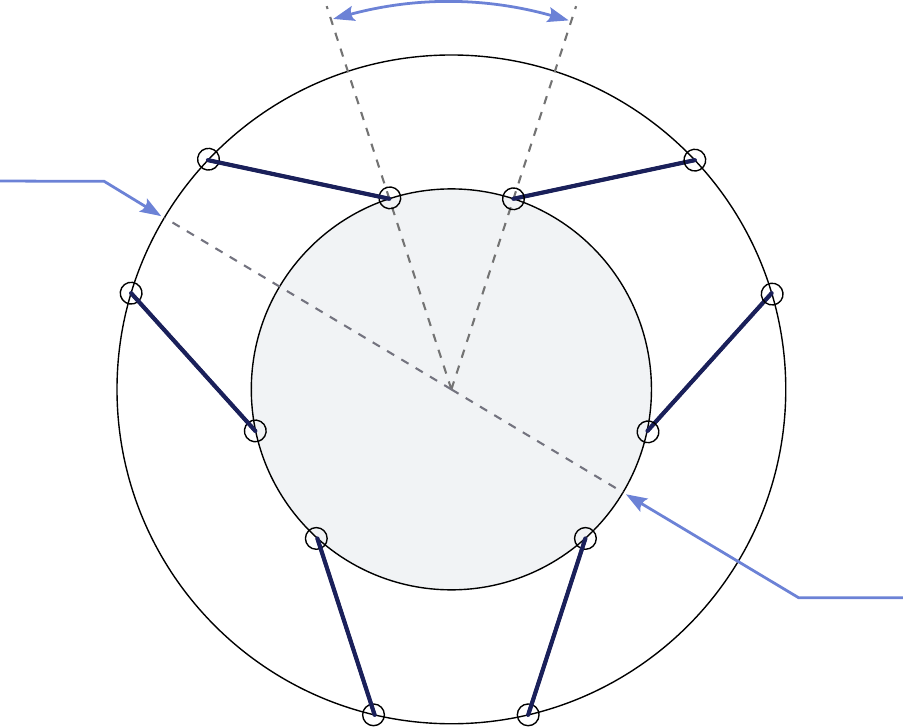}};
			\begin{scope}[ x={($0.01*(img.south east)$)}, y={($0.01*(img.north west)$)}, ]
	 		\begin{pgfonlayer}{annotation}
		 		\node[inner sep=2.0pt,anchor=south] at (6,75) {$r_b$};
		 		\node[inner sep=2.0pt,anchor=south] at (94,18) {$r_t$};
		 		\node[inner sep=2.0pt,anchor=south] at (50,100) {$\varphi_t$};
		 	\end{pgfonlayer}
	 		\end{scope}
		\end{tikzpicture}}
	\subcaptionbox{
		{Isometric view}
		\label{fig: geom parametrisaiton (b)}}[\boxw]{
		\begin{tikzpicture}
			\node[inner sep=0pt, anchor=south west] (img) at (0,0) {\includegraphics[width=\figw]{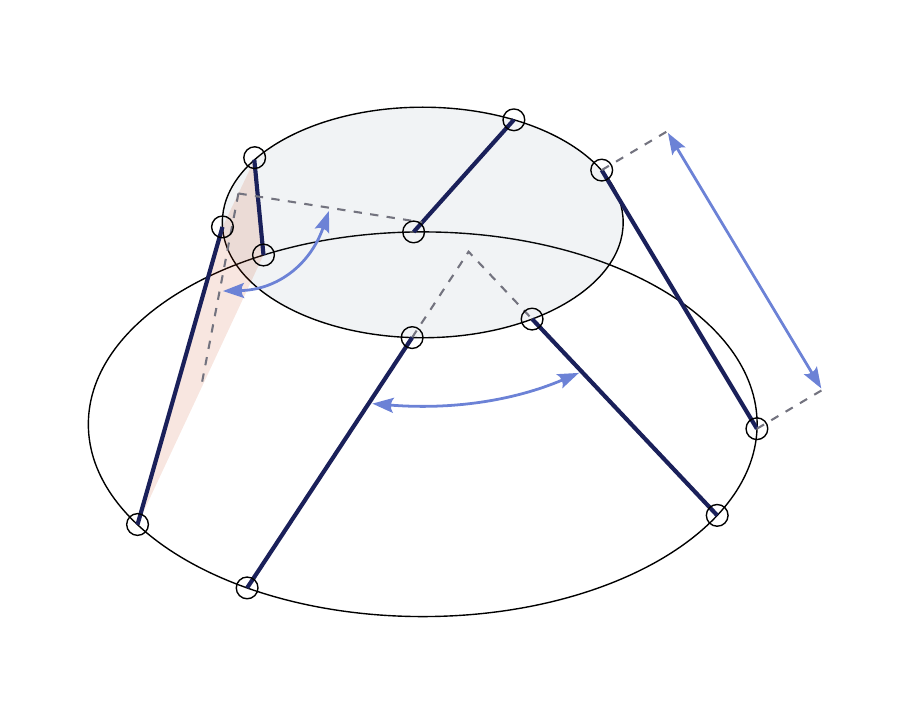}};
			\begin{scope}[ x={($0.01*(img.south east)$)}, y={($0.01*(img.north west)$)}, ]
	 		\begin{pgfonlayer}{annotation}
		 		\node[inner sep=3.0pt,anchor=north] at (53,44.5) {$\beta$};
			 	\node[inner sep=1.0pt,anchor=north west] at (33,62.5) {$\gamma$};
			 	\coordinate (A) at (82.5,63.5);
			 		\node[inner sep=2.5pt,anchor=south,rotate around={-59:(A)}] at (A) {$L$};
		 		\end{pgfonlayer}
	 		\end{scope}
		\end{tikzpicture}}
	\caption{Hexapod parametrisation. Any two of the lengths $L$, $r_t$, $r_b$ suffice to determine geometry}
	\label{fig: geom parametrisaiton}
\end{figure}
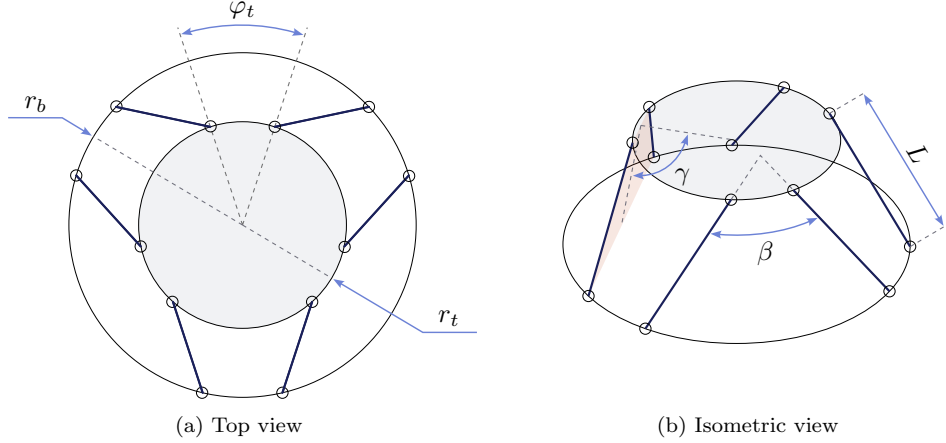

Various descriptions of a parallel manipulator's geometry can be accommodated by the model in \Cref{ssec: equations of motion,ssec: extension to arbitrary parallel manipulators,ssec: transmitted forces and moments}. Here, the parametrisation shown in \Cref{fig: geom parametrisaiton} is suggested for radially symmetric hexapods. It is based on quantities relevant in a top-level design context. Concretely, a strut length $L$, payload platform radius $r_t$, bipod link pair planar angle $\beta$ and angle $\gamma$ formed between a bipod's plane and the top platform at rest. To fully define the global geometry, an extra angle $\varphi_t$ that specifies the separation of two bipod pin joints in the plane of the moving platform is also introduced. For example, all planar and 3D structures studied in \Cref{sec: all-rotational limitations,sec: alternative BCs} have $\varphi_t = 0$, as the aforementioned pins coincide. On the other hand, a system made up of six upright links with uniform radial spacing would correspond to $\varphi_t = \pi/3$.

As a practical matter, the 3-1-3 Euler convention is suggested for obtaining $\m{R}_{i}$, since the constituent rotations are transparently related to $\beta$ and $\gamma$. In particular 
\begin{equation}\label{eq: 3-1-3 Rk}
	\m{R}_{i}(\psi_{i},\theta_{i},\phi_{i}) = 
	\begin{bmatrix}
		\cA_{i} 	& -\sA_{i} 	 & 0  	\\
		\sA_{i} 	& \cA_{i}	 & 0  	\\
		0   		& 0 	 	 & 1
	\end{bmatrix}
	\begin{bmatrix}
		1 			& 0 		& 0  	\\
		0 			& \cB_{i} 	& -\sB_{i}  \\
		0   		& \sB_{i}	& \cB_{i}
	\end{bmatrix} 
	\begin{bmatrix}
		\cC_{i} 	& -\sC_{i} 	 & 0  	\\
		\sC_{i} 	& \cC_{i}	 & 0  	\\
		0   		& 0 	 	 & 1
	\end{bmatrix}  
\end{equation}
When each link has a separate $\csys{C}_{i}$ aligned by $x_{i}$,
\begin{equation}
 	\psi_{i}  	= ({i}-1)\frac{\pi}{3}		 			,\quad 
 	\theta_{i} 	= \gamma - \pi							,\quad 
 	\phi_{i} 		= \frac{\pi}{2} \pm \frac{\beta}{2}
\end{equation}
with the the sign of $\beta$ determined by the parity of ${i}$. In the case of $\varphi_t = \beta = 0$, viz. \Cref{fig: hexapod displaced}, $\csys{C}_{i}$ are defined per bipod and $\phi_{i} = (\pi-\beta)/2$. Irrespectively of whether the hexapod is considered in terms of individual links or bipods, $\varphi_t$ is only necessary to determine $\vc{d}_{i}$ in \eqref{eq: T_ks} and does not affect $\m{R}_{i}$.

\section{Validation examples}
\label{sec: validation examples}
\subsection{Numerical implementation}
\label{ssec: numerical implementation}

For the purposes of this work, a MATLAB/C++ implementation adhering to \Cref{ssec: extension to arbitrary parallel manipulators} was developed. MATLAB's built-in ode113 variable-order method \cite{mat_shampine_97} was used for integration of \eqref{eq: state space def}. It exhibited better efficiency compared to common Runge-Kutta pairs. Convergence was quicker when using Tait-Bryan over proper Euler angles for $\m{\hat{R}}$, justifying the suggested 3-2-1 rotation sequence. Integrator relative tolerance of $10^{-9}$ and a maximum step size of $f^{-1}/10$ seconds were imposed.

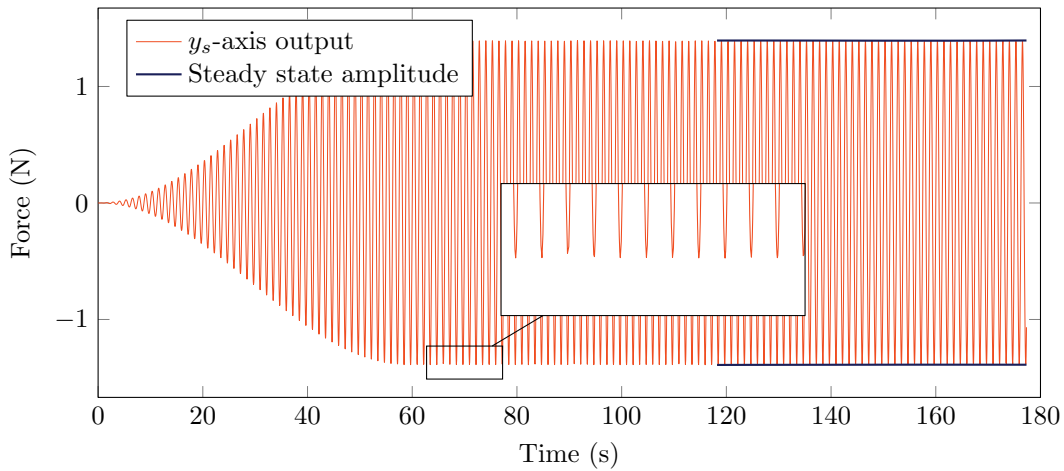
\begin{figure}[htb]
\centering
	\pgfplotsset{every axis/.append style=lin subplots,legend pos=north west} 
\begin{tikzpicture}[rectagular spy]
	\begin{axis}[xmin=0,xmax=180,xlabel={Time (s)},ylabel={Force (N)},grid=none,]
		\addplot[rc,ultra thin,line join=round] table {TikzData/TD-to-TF/FX-3.tsv};
			\addlegendentry{$y_s$-axis output}
		\addplot[thick,dbc] table {TikzData/TD-to-TF/FX-1.tsv};
		 	\addlegendentry{Steady state amplitude}
		\addplot[forget plot,thick,dbc] table {TikzData/TD-to-TF/FX-2.tsv};
		 	\addlegendentry{Lower envelope}
	 	\coordinate (spypoint) at (axis cs:70,-1.37); 
	 	\coordinate (spyviewer) at (axis cs:106,-.4); 
	\end{axis}
		\spy[width=4.0cm,height=1.75cm] on (spypoint) in node [fill=white] at (spyviewer);
\end{tikzpicture}
\caption{Estimation of $y_b - y_s$ input-to-output ratio for Hexapod $3$ at \SI{0.8}{\Hz}. Computed steady-state amplitude is $\SI{1.39}{\N} = \SI{2.87}{\dB}$. The excitation ramps up over $47.3$ cycles with $t_r = \SI{59.1}{\s}$}
\label{fig: TF from time signal}	
\end{figure}

Transfer functions were reconstructed from transient solutions as follows. A smoothly ramped-up sinusoidal excitation was applied to the structure at individual  payload DOFs and discrete frequencies. Each input signal $y(t)$ has the piecewise form
\begin{equation}
 y(t) = 
 \begin{dcases} 
  	\sin^2 (\pi t /(2t_r)) \sin(\omega t),	 & 0\leq t\leq t_r	 	\\
 	\sin(\omega t),							 & t > t_r	 
 \end{dcases}
\end{equation}
where $\sin^2 (\pi t /(2t_r))$ is monotonically increasing from 0 to 1 on $[0,t_r]$, with $dg/dt=0$ at 0 and $t_r$. As usual, $\omega = 2\pi f$ is the load circular frequency. The procedure aimed to avoid contamination of computed TFs by nonperiodic transient behaviour, e.g. see zoomed area of \Cref{fig: TF from time signal}. Variable ramp-up time was selected empirically and the transmitted amplitude extracted from the dynamic steady state. This method was found more accurate and efficient than taking a discrete Fourier transform of the output.

\subsection{Tested pin-slider hexapods}
\label{ssec: tested pin-slider hexapods}

\begin{figure}[htb]
	\centering
	\captionsetup[subfigure]{skip=1.0\subfigskip}
	\setl{boxw}{0.275\columnwidth}
	\setl{figw}{0.205\columnwidth}
	\subcaptionbox{
 		{Hexapod 1, cubic}
		\label{fig: various hexapods (a)}}[\boxw]{
		\includegraphics[width=\figw]{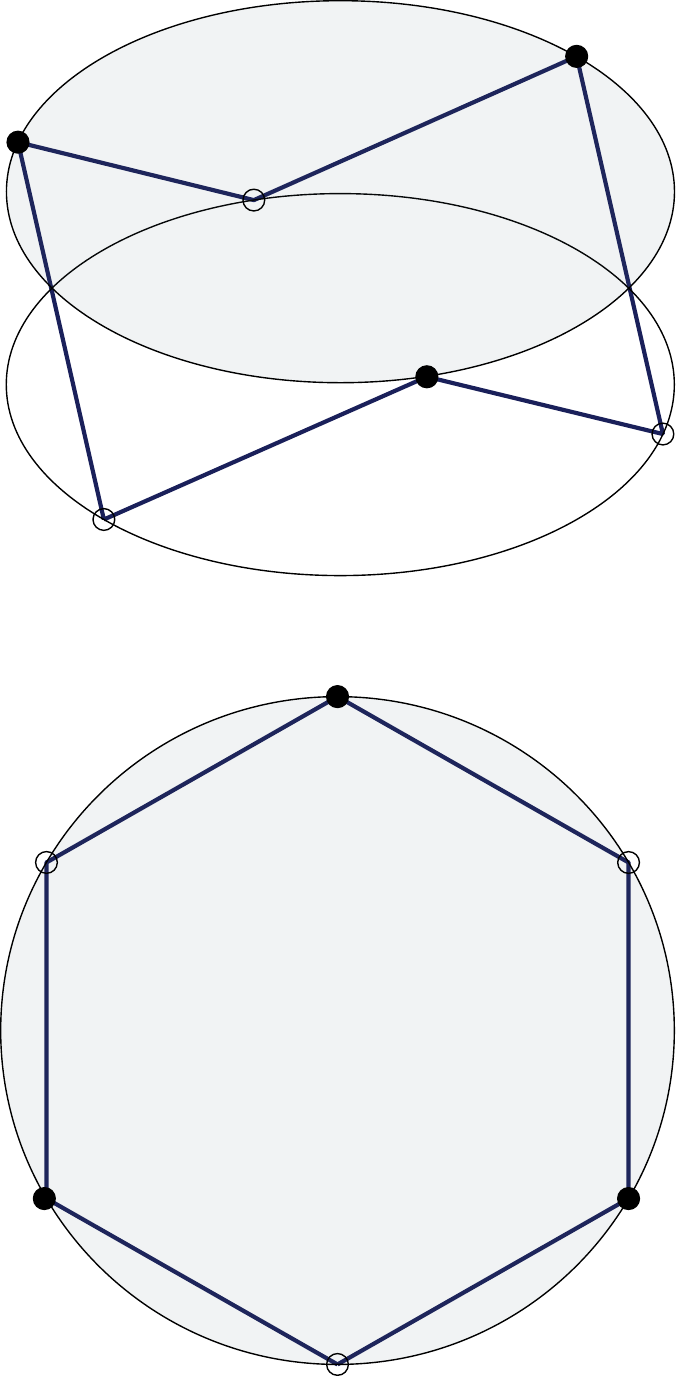}}
	\subcaptionbox{
 		{Hexapod 2, conic}
		\label{fig: various hexapods (b)}}[\boxw]{
		\includegraphics[width=\figw]{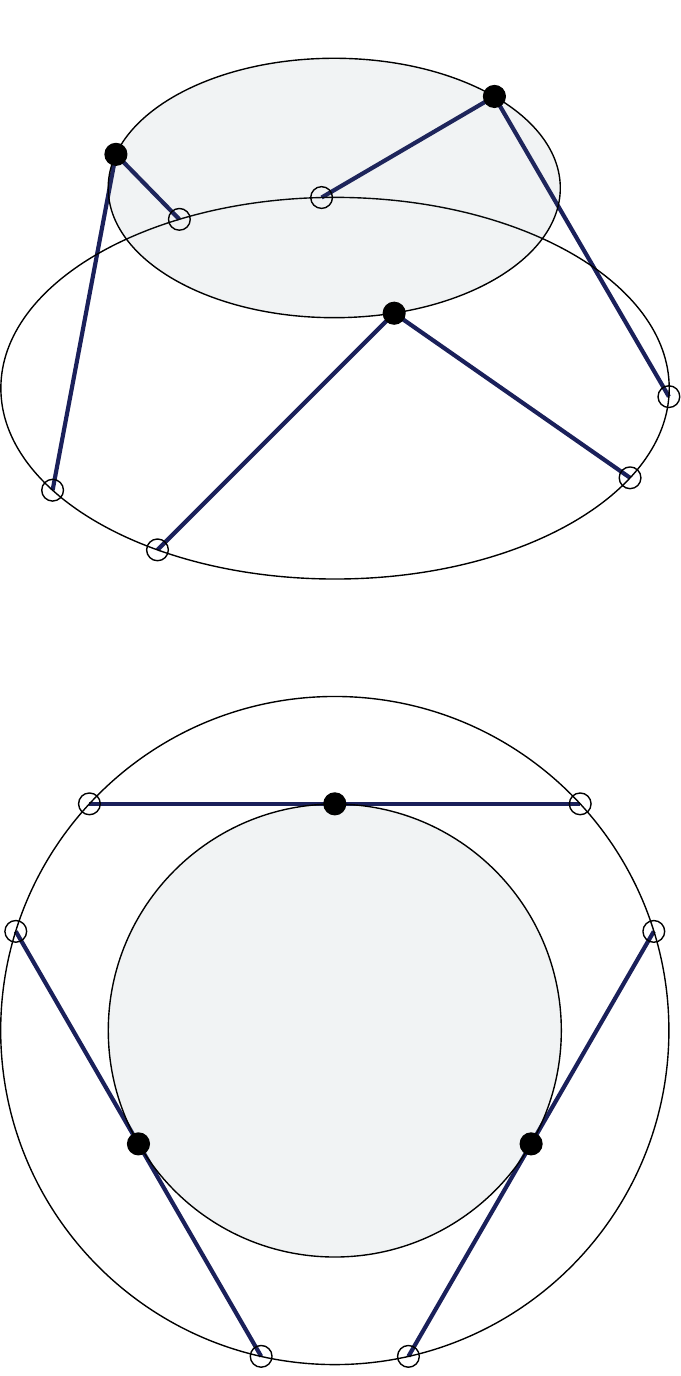}}
	\subcaptionbox{
 		{Hexapod 3, general}
		\label{fig: various hexapods (c)}}[\boxw]{
		\includegraphics[width=\figw]{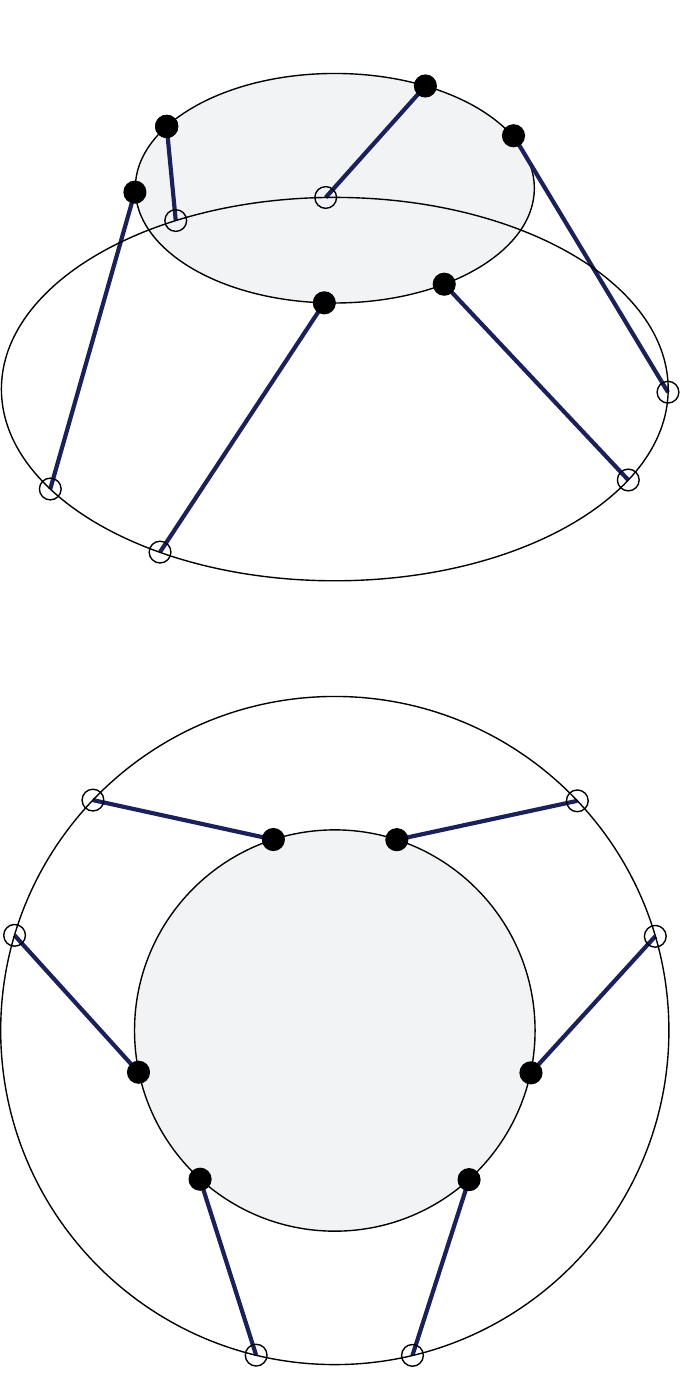}}
	\caption{Isometric view (top) and top view (bottom) of the systems defined by \Cref{tab: pin-slider hexapods}}
	\label{fig: various hexapods}
\end{figure}

To assess the viability of the proposed BC, the dynamic behaviour of three different pin-slider hexapods was investigated. Their geometries are defined in \Cref{tab: pin-slider hexapods} and illustrated in \Cref{fig: various hexapods}. Top platform radius $r_t$ was kept constant, as it would primarily be driven by the payload shape. Link properties are identical to the case study in \Cref{ssec: real struts} and are thus available in \Cref{tab: 2D pin-pin bipod}. Complementing data in \Cref{tab: 2D pin-pin bipod,tab: pin-slider hexapods}, the equipment is centrally placed, such that its CM lies on the axis of radial symmetry of each hexapod. The CM is positioned \SI{30}{\mm} below the moving platform's plane and its principal axes at rest coincide with $\csys{C}_s$, yielding $\m{\hat{R}}_0 = \m{E}$, thereby $\vc{u}_{i}^b = \vc{d}_{i} ~\forall {i}$. Finally, $\m{I}^b = \diag(0.7608,\, 0.7608,\, 0.48)$ \si{\kg\m\squared}. The ground constraint point $G$ is taken as the centre of the bottom platform. Note that \Cref{fig: plateau for different hexapods} does not correspond to pin-pin versions of the configurations analysed in this section. In \Cref{fig: various hexapods} the hexapods appear to have $r_b$ rather than $r_t$ fixed, but their scaling is not proportional.

\begin{table}[htb]
\centering
	\setl{tcw}{.1\columnwidth}
\begin{threeparttable}
\caption{Geometry of the hexapods in \Cref{fig: TFs hexapods diag}}
\label{tab: pin-slider hexapods}	
    \begin{tabular}{L{1.0\tcw}L{1.2\tcw}L{1.2\tcw}L{1.2\tcw}l}
	\toprule
			Property 	& 	Hexapod 1	 		& 	Hexapod 2	& Hexapod 3		& Unit 				\\
	\midrule
			$r_t$		& 0.245					&  	0.245		& 0.245			& \si{\m}			\\
			$\beta$		& $\pi/2$				& 	$\pi/2$		& $2\pi/5$ 		& \si{\radian}	\\
			$\gamma$	& $\arctan(\sqrt{2})$ 	& 	$\pi/2$ 	& $3\pi/5$ 		& \si{\radian}	\\
			$\varphi_t$	& 0						&  	0			& $\pi/6$		& \si{\radian}	\\
	\bottomrule
    \end{tabular}
\end{threeparttable}
\end{table}

Regarding test case selection, this section intends to demonstrate a wide range of systems with respect to the parametrisation from \Cref{ssec: hexapod geometry parametrisation}. For instance, the cubic configuration is well-studied for all-rotational joints and is characterised by pairwise orthogonal struts, as viewed from both platforms. It includes multiple intrinsic simplifications, in the sense of (theoretically) adjacent joints and \SI{90}{\degree} angles. Hexapod $3$, on the other hand, is a completely general Stewart platform. Hexapod $2$ represents an intermediate step, maintaining link orthogonality at moving platform joints and a vertical bipod orientation, but relaxing other constraints in comparison to the octahedral hexapod, such as platform radius to link length ratio. Since the BC proposed in this work inherently affects system dynamics, previous studies pertaining to pin-pin BCs are not applicable. In effect, analysing hexapods with varied geometrical features aims to provide preliminary data for future analysis and design, but not to find a universally preferable solution.

\subsection{Isolation performance results}
\label{ssec: isolation performance results}

While the governing equations are nonlinear, the small-workspace microvibration environment permits the payload-to-base disturbance transmission to be expressed in terms of a transfer function ${H}(f,\vc{g})$. Denote the elements of the vectors $\vc{f}_\text{in}^b$, $\vc{\uptau}_\text{in}^b$, $\vc{f}_\text{out}$ and $\vc{\upmu}_{\text{r},G}$ respectively by $F_{\text{in},j}^{b}$, $T_{\text{in},j}^{b}$, $F_{\text{out},q}$ and $M_{\text{r},G,q}$ for an axis $j$ in $\csys{C}_b$ and $q$ in $\csys{C}_s$. Then
\begin{equation}\label{eq: TF H(t,g)}
	\begin{bmatrix}
		F_{\text{out},x_s}  \\
		F_{\text{out},y_s}  \\
		F_{\text{out},z_s}  \\
		M_{\text{r},G,x_s}  \\
		M_{\text{r},G,y_s}  \\
		M_{\text{r},G,z_s} 
	\end{bmatrix} 
	\approx	
	\begin{bmatrix}
		{H}_{1,1} 	& {H}_{1,2} & {H}_{1,3} & {H}_{1,4} & {H}_{1,5} & {H}_{1,6} \\
		{H}_{2,1} 	& {H}_{2,2} & {H}_{2,3} & {H}_{2,4} & {H}_{2,5} & {H}_{2,6} \\
		0 			& 0 		& {H}_{3,3} & 0 		& 0 		& 0 		\\
		{H}_{4,1} 	& {H}_{4,2} & {H}_{4,3} & {H}_{4,4} & {H}_{4,5} & {H}_{4,6} \\
		{H}_{5,1} 	& {H}_{5,2} & {H}_{5,3} & {H}_{5,4} & {H}_{5,5} & {H}_{5,6} \\
		0 			& 0 		& {H}_{6,3} & 0 		& 0 		& {H}_{6,6} 
	\end{bmatrix} 
	\begin{bmatrix}
		F_{\text{in},x_b}^{b}  \\
		F_{\text{in},y_b}^{b}  \\
		F_{\text{in},z_b}^{b}  \\
		T_{\text{in},x_b}^{b}  \\
		T_{\text{in},y_b}^{b}  \\
		T_{\text{in},z_b}^{b} 
	\end{bmatrix}
\end{equation}
The zero entries arise from geometry and the nature of the slider BC. In addition, exploiting the rotational symmetry of the hexapods considered, ${H}_{1,1}={H}_{2,2}$, ${H}_{4,4}={H}_{5,5}$, ${H}_{2,4}={H}_{1,5}$, ${H}_{4,2}={H}_{5,1}$ and ${H}_{4,3}={H}_{5,3}$, up to minor deviations. Proper payload placement, for instance by aligning the primary mass CM with the geometric centre of the upper platform, may result in further elimination of cross-terms. However, such a study is beyond the scope of the current article. 

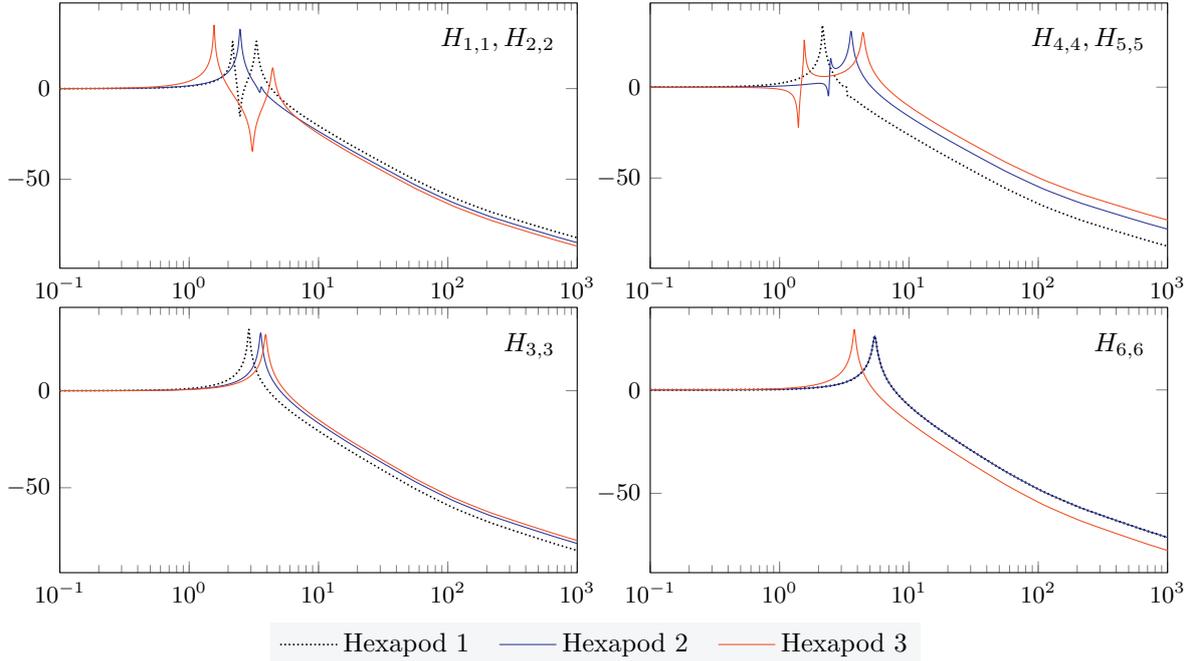
\begin{figure}[htb]
	\centering
	\pgfplotsset{every axis/.append style=matrix subplots,}
	\begin{tikzpicture}[ampersand replacement=\&,]
		\matrix[column sep=-6pt,]{
		\begin{axis}[	legend columns=-1,
						legend entries={Hexapod $1$, Hexapod $2$, Hexapod $3$,},
						legend to name=lgd,
						label style={font=\scriptsize},
						legend style=matrix subplots lgd,
						]
			\addplot[hexapod 1] table {TikzData/TFs/H1_1-1.tsv};
			\addplot[hexapod 2] table {TikzData/TFs/H2_1-1.tsv};
			\addplot[hexapod 3] table {TikzData/TFs/H3_1-1.tsv};
			\node[anchor=north east] at (rel axis cs:.975,.94) {${H}_{1,1}, {H}_{2,2}$};
		\end{axis}
		\&
		\begin{axis}[]
			\addplot[hexapod 1] table {TikzData/TFs/H1_4-4.tsv};
			\addplot[hexapod 2] table {TikzData/TFs/H2_4-4.tsv};
			\addplot[hexapod 3] table {TikzData/TFs/H3_4-4.tsv};
			\node[anchor=north east] at (rel axis cs:.975,.94) {${H}_{4,4}, {H}_{5,5}$};
		\end{axis}
		\\
		\begin{axis}[]
			\addplot[hexapod 1] table {TikzData/TFs/H1_3-3.tsv};
			\addplot[hexapod 2] table {TikzData/TFs/H2_3-3.tsv};
			\addplot[hexapod 3] table {TikzData/TFs/H3_3-3.tsv};
			\node[anchor=north east] at (rel axis cs:.975,.94) {${H}_{3,3}$};
		\end{axis}
		\&
		\begin{axis}[]
			\addplot[hexapod 1,thick] table {TikzData/TFs/H1_6-6.tsv};
			\addplot[hexapod 2] table {TikzData/TFs/H2_6-6.tsv};
		 	\addplot[hexapod 3] table {TikzData/TFs/H3_6-6.tsv};
			\node[anchor=north east] at (rel axis cs:.975,.94) {${H}_{6,6}$ };
		\end{axis}
		\\};
	\end{tikzpicture}
	\ref{lgd} 
	\caption{Diagonal TF matrix elements for the pin-slider hexapods, magnitude (\si{\dB}) vs. frequency (\si{\Hz}). Note that ${H}_{6,6}$ for the cubic and conic hexapod coincide}
	\label{fig: TFs hexapods diag}	
\end{figure}

The $H_{j,j}$ diagonal transfer function terms of \eqref{eq: TF H(t,g)} were obtained according to \Cref{ssec: numerical implementation} and are reported in \Cref{fig: TFs hexapods diag}. The damped pin-slider systems do not exhibit an attenuation plateau, in agreement with the preliminary Simulink and ADAMS results from \Cref{ssec: pin-slider numerical tests}. A roll-off slope of \SI{-20}{\dB\per\dec} is observed at high frequency, in line with expectations. Hexapod 2, herein referred to as conic, demonstrates favourable dynamics, owed to the closest spacing of payload modes, with the exception of vertical axis rotation (see \Cref{tab: pin-slider hexapod NFs}). Note that the two in-plane modes consist of inherently coupled lateral translation and rotation. Nevertheless, for the conic configuration one of their natural frequencies coincides with the vertical translation mode, bringing it closer to dynamic isotropy than its counterparts. 

\begin{table}[htb]
	\centering
	\setl{tcw}{.1\columnwidth}
	\begin{threeparttable}
		\caption{Payload rigid body motion natural frequencies in \si{\Hz}}
		\label{tab: pin-slider hexapod NFs}	
		\begin{tabular}{L{1.2\tcw}L{1.0\tcw}L{1.2\tcw}L{1.2\tcw}l}
			\toprule
			Mode 					& Axes 			& 	Hexapod 1	& 	Hexapod 2	& Hexapod 3		\\
			\midrule
			In-plane $1$			& $x_s$, $y_s$	& 2.17			&  	2.46		&	1.55 		\\
			In-plane $2$			& $x_s$, $y_s$	& 3.31			& 	3.57		& 	4.42 		\\
			Translation				& $z_s$			& 2.90 			& 	3.57		&	3.90 		\\
			Rotation				& $z_s$			& 5.46			&  	5.46		& 	3.79		\\
			\bottomrule
		\end{tabular}
	\end{threeparttable}
\end{table}

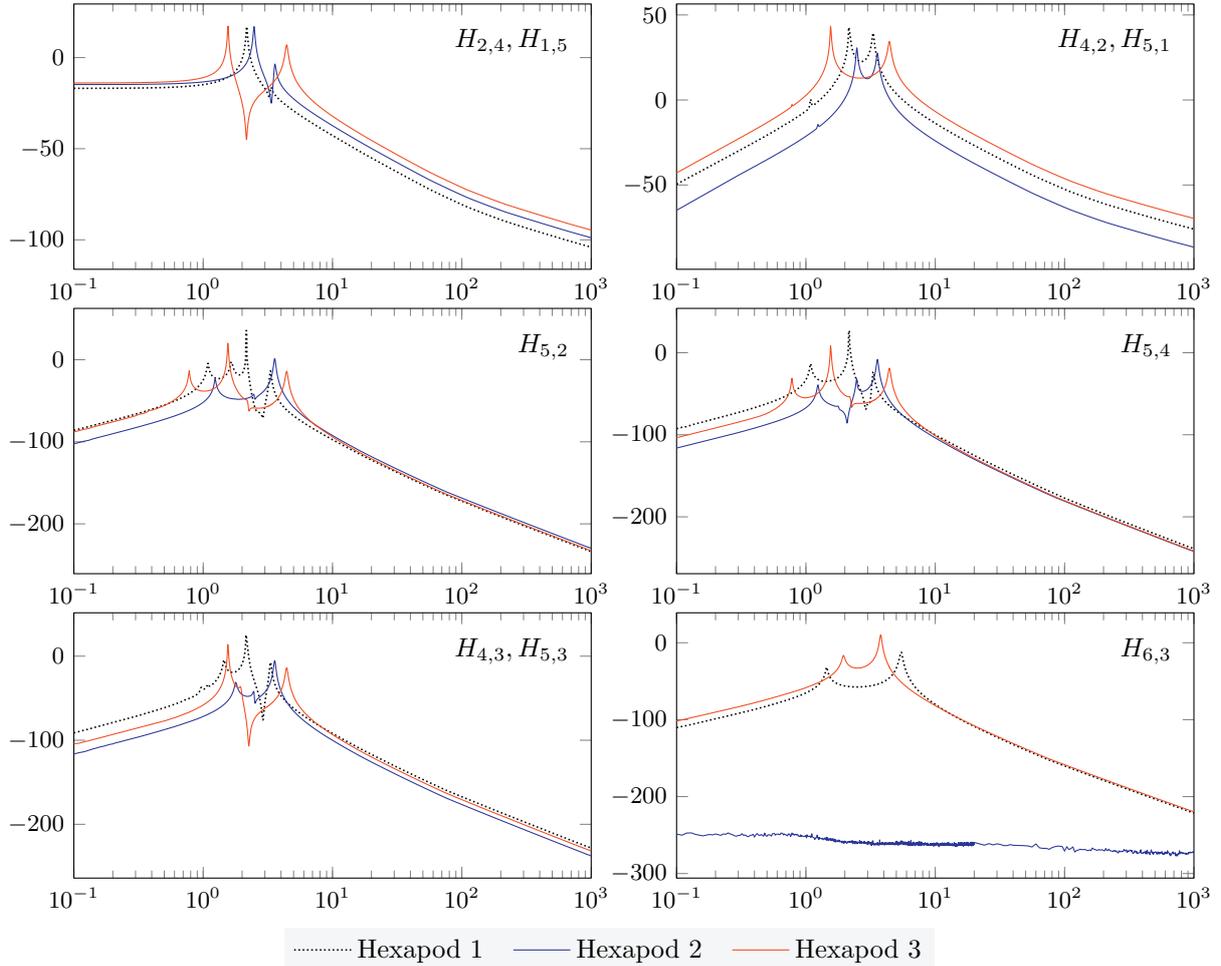
\begin{figure}[htb]
	\centering
	\pgfplotsset{every axis/.append style=matrix subplots,}
	\begin{tikzpicture}[ampersand replacement=\&,]
		\matrix[column sep=-6pt,]{
		\begin{axis}[	legend columns=-1,
						legend entries={Hexapod $1$, Hexapod $2$, Hexapod $3$,},
						legend to name=lgd,
						label style={font=\scriptsize},
						legend style=matrix subplots lgd,
						]
			\addplot[hexapod 1] table {TikzData/TFs/H1_2-4.tsv};
			\addplot[hexapod 2] table {TikzData/TFs/H2_2-4.tsv};
			\addplot[hexapod 3] table {TikzData/TFs/H3_1-5.tsv};
			\node[anchor=north east] at (rel axis cs:.975,.94) {${H}_{2,4}, {H}_{1,5}$};
		\end{axis}
		\&
		\begin{axis}[]
			\addplot[hexapod 1] table {TikzData/TFs/H1_4-2.tsv};
			\addplot[hexapod 2] table {TikzData/TFs/H2_4-2.tsv};
			\addplot[hexapod 3] table {TikzData/TFs/H3_4-2.tsv};
			\node[anchor=north east] at (rel axis cs:.975,.94) {${H}_{4,2}, {H}_{5,1}$};
		\end{axis}
		\\
		\begin{axis}[]
			\addplot[hexapod 1] table {TikzData/TFs/H1_5-2.tsv};
			\addplot[hexapod 2] table {TikzData/TFs/H2_5-2.tsv};
			\addplot[hexapod 3] table {TikzData/TFs/H3_5-2.tsv};
			\node[anchor=north east] at (rel axis cs:.975,.94) {${H}_{5,2}$};
		\end{axis}
		\&
		\begin{axis}[]
			\addplot[hexapod 1] table {TikzData/TFs/H1_5-4.tsv};
			\addplot[hexapod 2] table {TikzData/TFs/H2_5-4.tsv};
			\addplot[hexapod 3] table {TikzData/TFs/H3_5-4.tsv};
			\node[anchor=north east] at (rel axis cs:.975,.94) {${H}_{5,4}$};
		\end{axis}
		\\
		\begin{axis}[]
			\addplot[hexapod 1] table {TikzData/TFs/H1_4-3.tsv};
			\addplot[hexapod 2] table {TikzData/TFs/H2_5-3.tsv};
			\addplot[hexapod 3] table {TikzData/TFs/H3_5-3.tsv};
			\node[anchor=north east] at (rel axis cs:.975,.94) {${H}_{4,3}, {H}_{5,3}$};
		\end{axis}
		\&
		\begin{axis}[]
			\addplot[hexapod 1] table {TikzData/TFs/H1_6-3.tsv};
			\addplot[hexapod 2] table {TikzData/TFs/H2_6-3.tsv};
			\addplot[hexapod 3] table {TikzData/TFs/H3_6-3.tsv};
			\node[anchor=north east] at (rel axis cs:.975,.94) {${H}_{6,3}$};
		\end{axis}
		\\};
	\end{tikzpicture}
	\ref{lgd} 
	\caption{Off-diagonal TF matrix elements for the pin-slider hexapods, magnitude (\si{\dB}) vs. frequency (\si{\Hz}). For the conic configuration, $H_{6,3}=0$ as the plotted response equals the machine epsilon}
	\label{fig: TFs hexapods off-diag}	
\end{figure}

A collection of off-diagonal transfer function terms is also presented in \Cref{fig: TFs hexapods off-diag}. In general, the strongest excitation force - constraint moment and excitation torque - constraint force coupling occurs in-plane, viz. ${H}_{2,4}$, ${H}_{1,5}$, ${H}_{4,2}$ and ${H}_{5,1}$. The aforesaid TFs are characterised by a high-frequency isolation of \SI{-20}{\dB\per\dec}. The remaining off-diagonal terms of $H$ have a roll-off slope of \SI{-60}{\dB\per\dec}. Attention is drawn to the case of vertical torque - vertical constraint force. In particular, the force transmitted by the conic configuration is zero, since it falls within the machine epsilon region. When the bipods are not upright, that is, $\gamma \neq \pi/2$, coupling between $z_b$ translation and rotation about $z_s$ occurs, giving rise to $H_{6,3}\neq 0$. While potentially counterintuitive, this is indeed the correct behaviour and can be validated by a more careful examination of the motion constraint imposed by the strut base rollers.

Overall, the analytical model tests support intuition and affirm the benefit of the proposed slider BC for arbitrarily dimensioned parallel manipulators. Although the purpose of the comparison in this section was not to assess the viability of different strut arrangements, it is worth pointing out that the conic hexapod was closest to dynamic isotropy. It was also the only configuration exhibiting a zero $z_b$ torque to vertical ground force cross-contamination, while simultaneously presenting the lowest cross-term TF peaks. From an engineering perspective, this geometry is desirable because changing payload platform dimensions does not necessitate bipod redesign. In conclusion, the conducted study indicates that the most promising pin-slider hexapod setup in terms of high-frequency attenuation and practicality appears to be a construction of three vertical bipods, each having an orthogonal pair of struts. Nonetheless, finding a globally optimal geometry with respect to given performance targets is outside the scope of the article.

Pin-slider parallel manipulators having a number of links $n\neq 6$ and/or lacking rotational symmetry generally exhibit the same improved high-frequency behaviour as their symmetric Gough-Stewart counterparts. However, the latter are prevalent in practice, which is why the case studies considered here have focused on them. In terms of future design guidelines for isolation platforms, the new BC indeed proves beneficial. It circumvents the attenuation plateau conferred in \Cref{ssec: real struts} and therefore enables passive or semi-active isolators to be designed for broadband attenuation purposes. This could prove vital for space applications, where mitigation of high-frequency disturbances clashes with mass and complexity requirements not being met by active devices.

\section{Potential slider design}
\label{sec: planar joint}

A common concern when dealing with microvibrations is the feasibility of traditional mechanism design approaches. Low-level rattling or even friction within the joints could severely affect the attenuation efficiency of an isolator or compromise the transmission path of a support device. To that end, multi-part joints such as ball bearings or S-joints are often replaced by devices made of continuous elements, like flexures or rods. A linear stiffness approximation in the $6$ spatial DOFs is justified due to the small-displacement regime implied by the microvibration environment. Examples of flexure-based U-joint substitutes can be seen in \cite{preumont:single_stage_strut, hauge:six_axis_vibr_isolation}. In order to reproduce the planar joint suggested for a link-base connection in this paper, several ideas have been evaluated and the most promising is reported below. The basic concept exploits carbon fibers' favourable combination of high Young's modulus, ranging from \SI{250}{\GPa} to \SI{600}{\GPa}, and virtually zero bending and shear stiffness. This refers to the 'naked' fibres only, that is, not within a polymer matrix. Indeed, shear stiffness of unidirectional composite panels is dictated almost entirely by resin mechanical properties. However, in the proposed joint, the resin would not be necessary as the carbon fibres would be pre-tensioned by a spring. Provided that the nominal compressive force is smaller than the preload, the fibres would only operate in tension. Due to the very small anticipated excitations, the preload can remain relatively low, in the order of tens of Newtons. 

\begin{figure}[htb]
\centering	
	\captionsetup[subfigure]{skip=\subfigskip}
	\setl{boxw}{0.36\columnwidth}	
	\setl{figh}{0.34\columnwidth} 
\subcaptionbox{
	{CAD model}
	\label{fig: slider (a)}}[\boxw]{
	\includegraphics[height=\figh]{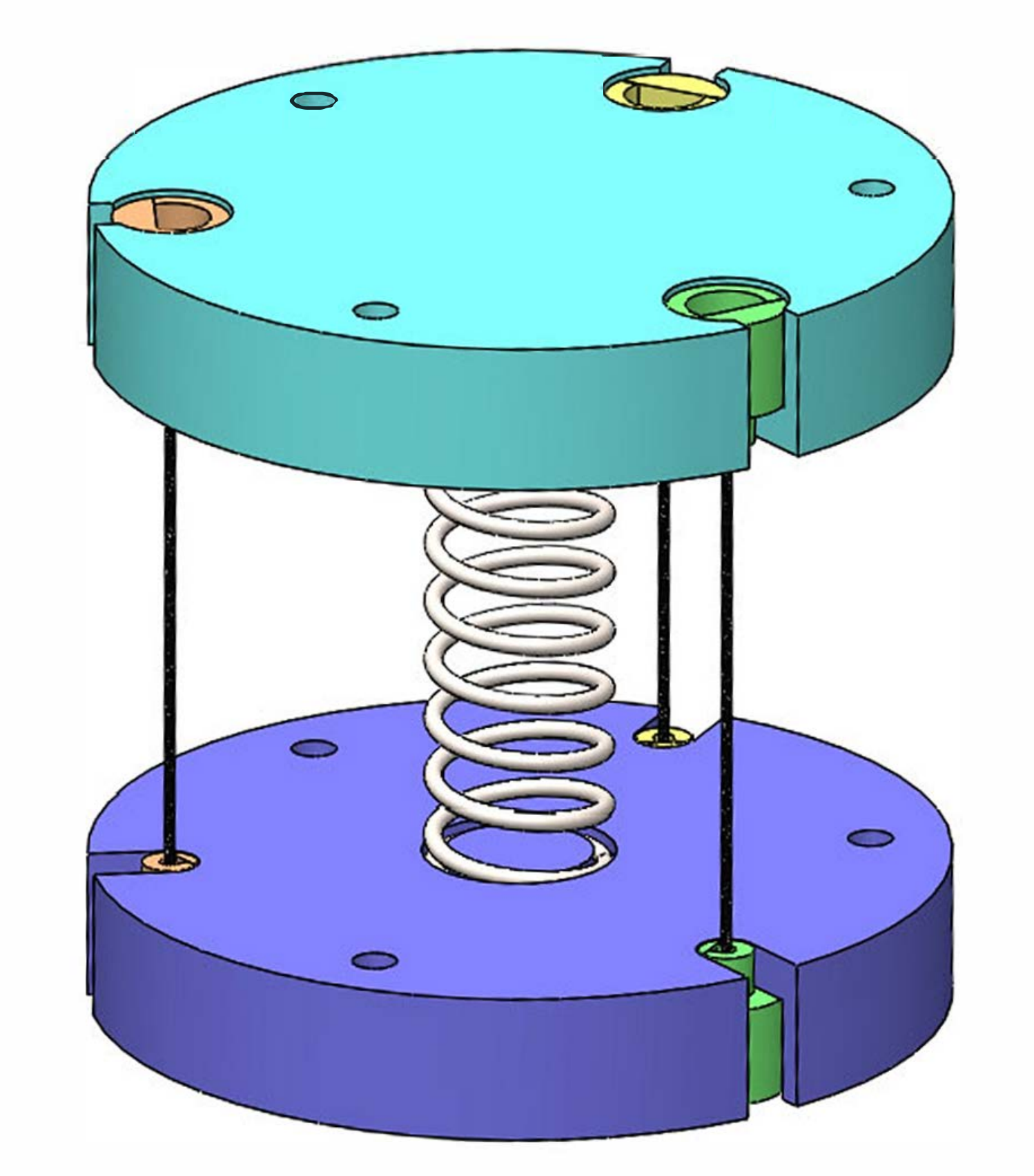}}
\subcaptionbox{
	{Prototype built in-house}
	\label{fig: slider (b)}}[\boxw]{
	\includegraphics[height=\figh]{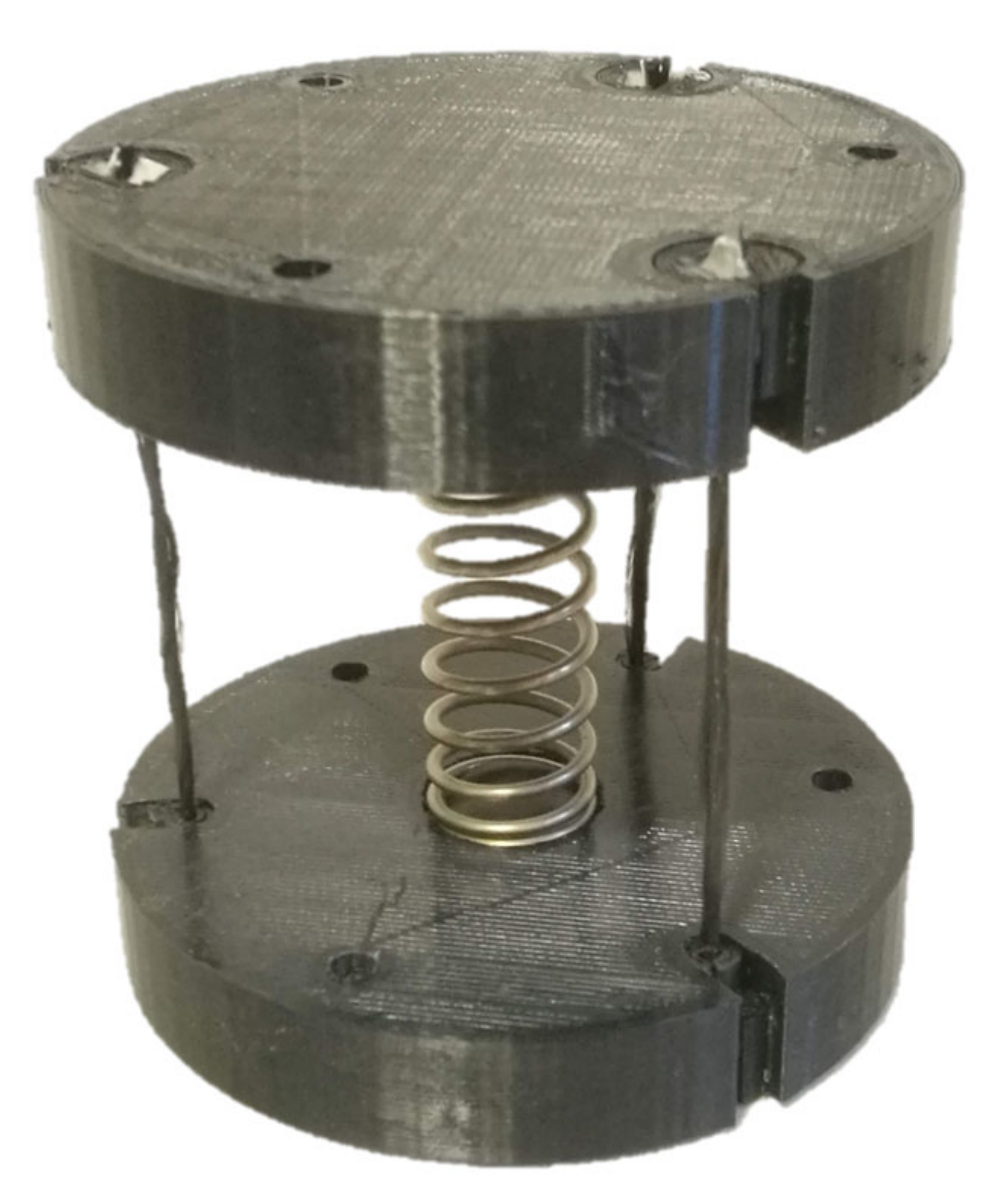}}
\caption{Proposed slider solution, relying on carbon fibre tows kept in tension by a preload spring}
\label{fig: slider}
\end{figure}

An example of the proposed planar joint can be seen in \Cref{fig: slider}. It is made of three tows of 'naked' carbon fibres adhering to a radially symmetric placement with respect to a preloaded spring that ensures they are in tension. This arrangement keeps the joint isostatic. A prototype was built in house using carbon fibre reel and a stainless steel spring with an axial stiffness of \SI{2036}{\N\per\m} and free length on \SI{63}{\mm}. The fully assembled mechanism is displayed in \Cref{fig: slider (b)}. It has dimensions of \SI{86}{\mm} in diameter and \SI{85}{\mm} in height, with the fibre tows being \SI{54}{\mm} long and preloaded with a \SI{12}{\newton} total force. Two sets of experiments were performed to estimate the stiffness of the joint. 

\begin{figure}[htb]
\centering
	\captionsetup[subfigure]{skip=.5\subfigskip}
	\setl{boxw}{\columnwidth}
	\setl{figw}{0.75\columnwidth}
\subcaptionbox{
	\label{fig: slider test (a)}}[\boxw]{
	\includegraphics[width=\figw]{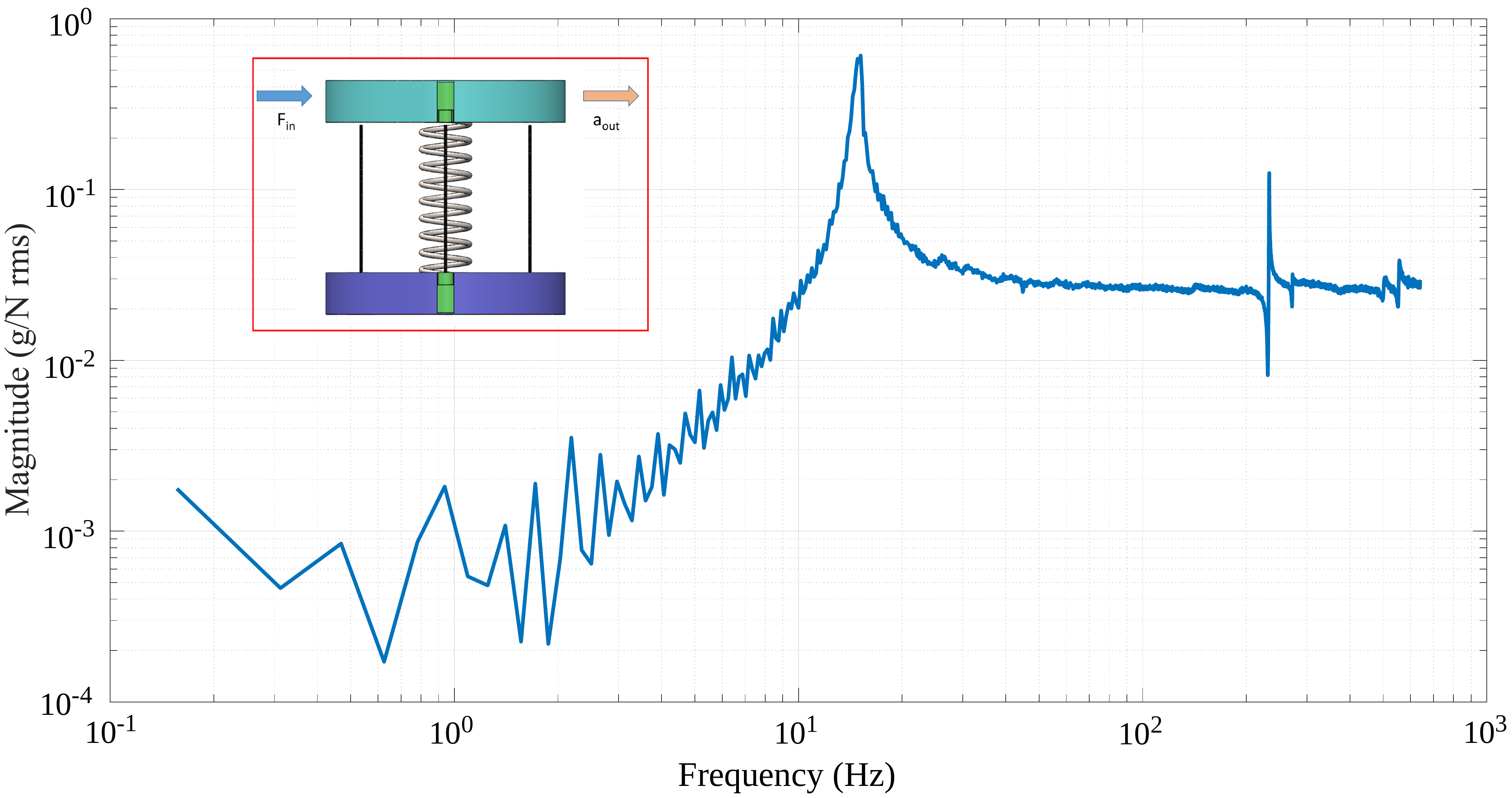}}
\subcaptionbox{
	\label{fig: slider test (b)}}[\boxw]{
	\includegraphics[width=\figw]{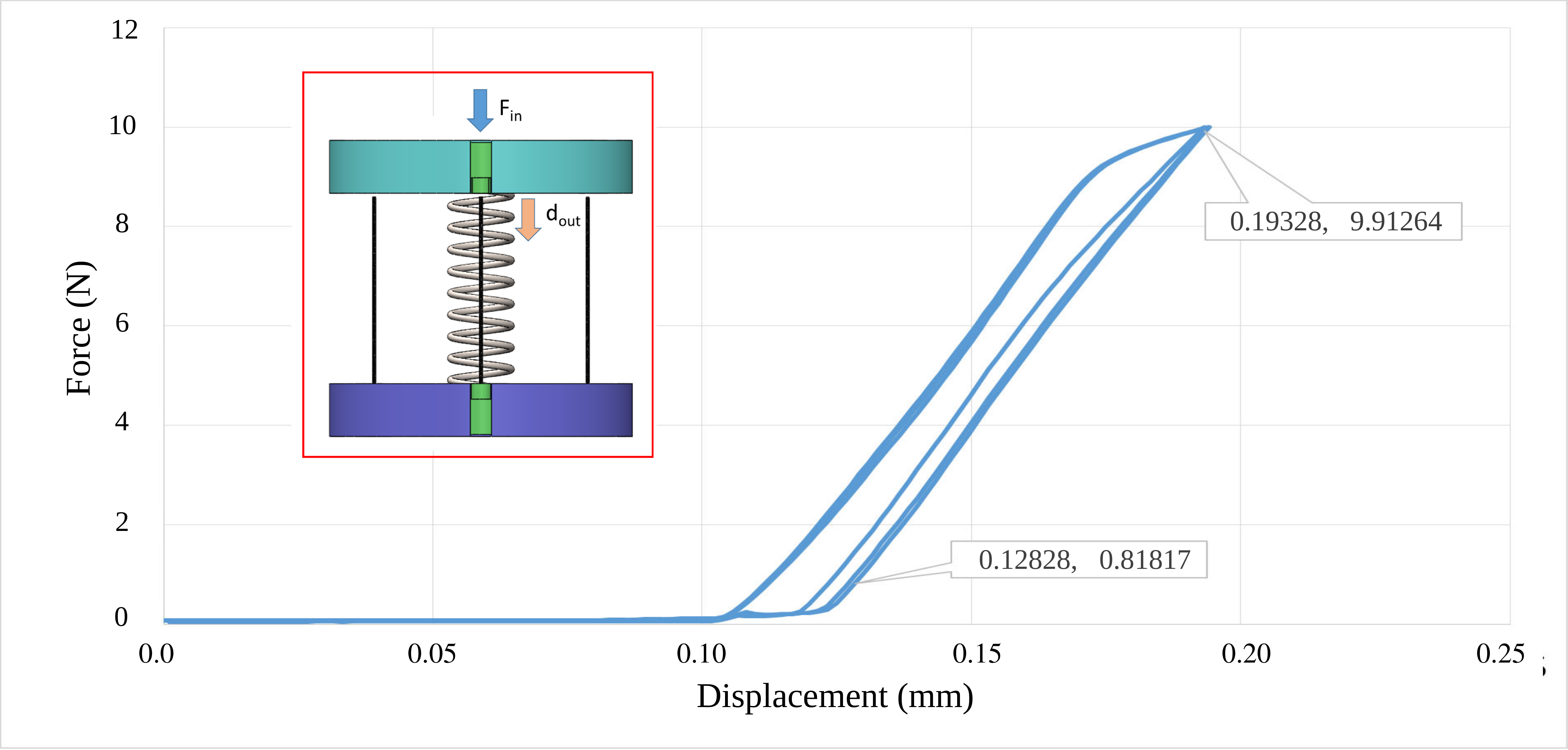}}
\caption{Tests to assess slider stiffness: \subref{fig: slider test (a)} top platform lateral excitation - lateral acceleration TF recorded from an impact test, \subref{fig: slider test (b)} force-displacement plot from a tension-compression machine}
\label{fig: slider test}
\end{figure}

An impact hammer test was used to assess the shear stiffness of the joint. The bottom platform was clamped and a lateral force was introduced to the joint by impacting along the geometric centre of the top platform. The acceleration along the conforming direction was measured with two accelerometers. The combined TF from the two sensors is plotted in \Cref{fig: slider test (a)}. From the figure, the lateral mode at \SI{15.2}{\Hz} and the bending mode at \SI{220}{\Hz} are clearly distinguished. The mass of the platform plus two the accelerometers was \SI{0.055}{\kg}, resulting in a shear stiffness of $\approx \SI{500}{\N\per\m}$. The value falls within the expected range, considering that the spring in the preloaded condition presents a lateral stiffness of approximately \SI{300}{\N\per\m}, as measured with a compression-tension machine. The additional contribution due to the three tows amounts to \SI{220}{\N\per\m}. This estimation is obtained by taking the boundary conditions of the carbon fibre strands to be pin at the bottom and roller at the top and assuming small displacements. The stated value is then easily confirmed by considering a moment equilibrium about the bottom pin and the aforesaid planar joint dimensions, namely strand length. It is important to note that the theoretical joint transverse stiffness depends solely on the preload force, tow length and shear stiffness of the spring. 

\begin{table}[H]
\centering
	\setl{tcw}{.1\columnwidth}
	\sisetup{
			 	table-number-alignment=left, 	
				table-format=3e+3,				
				table-column-width=.9\tcw, 		
			}
\begin{threeparttable}
\caption{Planar joint prototype equivalent stiffness}
\label{tab:slider_stiffness}	
    \begin{tabular}{L{2.2\tcw}S[]l}
	\toprule
    	Direction 	& \multicolumn{1}{l}{Value} 	& Unit 		\\
	\midrule
	 	Axial, measured			& 140.0		& \si{\kN\per\m} 	 		\\
		Shear, measured 		& 0.5		& \si{\kN\per\m}			\\
		Bending, computed		& 83.0		& \si{\N\m\per\radian}	\\
		Torsion, computed  		& 1.83		& \si{\N\m\per\radian}	\\
	\bottomrule
    \end{tabular}
\end{threeparttable}
\end{table}

The second test performed was aimed at evaluating the axial stiffness of the slider. A compression-tension machine was used and the resultant force-displacement plot can be seen in \Cref{fig: slider test (b)}. Apart from a narrow hysteresis cycle, it transpires that the axial stiffness is linear at least up to the maximum compression of \SI{10}{\newton} that was applied and equals approximately \SI{140}{\kilo\N\per\m}. The remaining properties of the slider reported in \Cref{tab:slider_stiffness} have been computed through the measured ones. In particular, the bending stiffness is directly proportional to the axial stiffness and the side length of the equilateral triangle formed by the carbon fibre attachments. Similarly, the torsional stiffness is proportional to the shear stiffness and the circumcircle radius of the triangle. The contribution of the spring is negligible, due to the proximity with the axis of rotation. The prototype's large ratios of relevant stiffness pairs can be already appreciated, even though further analysis would be necessary to optimise the joint. Specifically, it would be desirable to obtain higher in-plane and torsional compliance while maintaining the relative rigidity in the other directions, thus better replicating the theoretical slider BC.

\section{Conclusions}

The influence of different strut boundary conditions on the isolation properties of parallel manipulators was studied in the present article. While passive devices are preferable for space missions from complexity and reliability point of view, it was shown that satisfactory high-frequency performance cannot be achieved by the state-of-the-art all-rotational joint systems. Alternative boundary conditions have been explored and a slider replacement for the usual pin connection between links and fixed platform is proposed. Equations of motion for an arbitrary pin-slider parallel manipulator were derived under a minimal set of assumptions. Numerical tests confirm that disturbance transfer functions do not exhibit a high-frequency plateau. In fact, they were found to behave similarly to ones ascribed to pin-pin platforms employing unrealistic massless strut models. The theoretically ideal attenuation of \SI{-40}{\dB\per\dec} is observed after the last payload rigid motion mode in the absence of damping and \SI{-20}{\dB\per\dec} with dashpots. A pin-slider hexapod configuration with favourable dynamics is suggested. Finally, a prototype planar joint that could reproduce the proposed lateral slider condition is outlined. It showed promising results in terms of translational to shear stiffness ratio, which exceeded $2$ orders of magnitude, as well as bending to torsional stiffness, which was close to $50$. In summary, this work outlines a new link boundary condition and corresponding planar joint design that enable the development of high-performance passive and semi-active hexapod platforms for microvibration suppression. 


\bibliographystyle{ieeetr}
\bibliography{bib_MMT.bib}


\appendix
\section{Pin-slider equations derivtion}\label{sec: apx 3D deriv}

The 2\textsuperscript{nd} order derivatives appearing in $\vc{f}_\text{r,tot}$ can be separated and recast into matrix-vector product form: 
\begin{equation}\label{eq: f_r,tot sparation}
\begin{aligned}
 \vc{f}_{P,\text{tot}}  
   &= 
 \sum_{i}\nolimits \m{R}_{i} \vc{f}_{P_{i}}^{i}  
\\ &= 
 -\sum_{i}\nolimits \m{R}_{i}  \big(  {k}\m{P}\vc{u}_{i}^{i}  +  {c}\m{P}\dot{\vc{u}}_{i}^{i}  +  m_s\m{S}\ddot{\vc{u}}_{i}^{i}  \big)
\\ &= 
 -\sum_{i}\nolimits \m{R}_{i}\m{P}  \big( {k}\vc{u}_{i}^{i}  +  {c}\dot{\vc{u}}_{i}^{i}  \big)
 -\sum_{i}\nolimits m_s \underbracket[.24ex]{\m{R}_{i}\m{S}\mt{R}_{i}}_{\m{\bar{S}}_{i}} \Big( \ddot{\hat{\vc{d}}}  +   \m{\hat{R}}\skewt{\dot{\vc{\omega}}^b}\vc{u}_{i}^b  +  \m{\hat{R}} \skewt{\vc{\omega}^b}^2\vc{u}_{i}^b  \Big)
\\ &= 
 -\sum_{i}\nolimits 
  \underbracket[.24ex]{ \Big[ \m{R}_{i}\m{P}  \big( {k}\vc{u}_{i}^{i}  +  {c}\dot{\vc{u}}_{i}^{i}  \big)
  + m_s \m{\bar{S}}_{i}\m{\hat{R}} \skewt{\vc{\omega}^b}^2\vc{u}_{i}^b	\Big]
  }_{\vc{\bar{f}}_{P,{i}}}
 -m_s \sum_{i}\nolimits \m{\bar{S}}_{i} \ddot{\hat{\vc{d}}}  
 +m_s \underbracket[.24ex]{\sum_{i}\nolimits \m{\bar{S}}_{i}\m{\hat{R}}\skewt{\vc{u}_{i}^b}}_{\m{\bar{D}}} \dot{\vc{\omega}}^b
\\ &= 
 - m_s  \sum_{i}\nolimits \m{\bar{S}}_{i} \ddot{\hat{\vc{d}}}  +  m_s \m{\bar{D}}\dot{\vc{\omega}}^b   
 - \sum_{i}\nolimits \vc{\bar{f}}_{P,{i}}
\end{aligned}
\end{equation}
Here, $\sum_{i}\nolimits \vc{\bar{f}}_{P,{i}}$ is a nonlinear function of the state vector. Rearrangement for $\m{\bar{D}}$ exploits basic properties of the skew-symmetric operator. An analogous procedure for the $\csys{C}_b$ reaction torques yields
\begin{equation}\label{eq: tau_r,tot^b sparation}
\begin{aligned}
 \vc{\uptau}^b_{P,\text{tot}} 
   &= 
 \sum_{i}\nolimits \skewt{\vc{u}_{i}^b} \mt{\hat{R}} \m{R}_{i} \vc{f}_{P_{i}}^{i}
\\ &= 
 -\sum_{i}\nolimits \skewt{\vc{u}_{i}^b} \mt{\hat{R}} \vc{\bar{f}}_{P,{i}}
 -m_s \underbracket[.24ex]{\sum_{i}\nolimits \skewt{\vc{u}_{i}^b} \mt{\hat{R}}\m{\bar{S}}_{i}}_{-\mt{\bar{D}}} \ddot{\hat{\vc{d}}}
 +m_s \underbracket[.24ex]{ \sum_{i}\nolimits \skewt{\vc{u}_{i}^b} \mt{\hat{R}}\m{\bar{S}}_{i}\m{\hat{R}}  \skewt{\vc{u}_{i}^b}}_{\m{\bar{U}}^b}
  \dot{\vc{\omega}}^b
\\ &= 
 m_s \mt{\bar{D}} \ddot{\hat{\vc{d}}} 
 +m_s \m{\bar{U}}^b \dot{\vc{\omega}}^b
 -\sum_{i}\nolimits \skewt{\vc{u}_{i}^b} \mt{\hat{R}} \vc{\bar{f}}_{P,{i}}
\end{aligned}
\end{equation}
Notice that $\displaystyle\sum_{i}\nolimits \skewt{\vc{u}_{i}^b} \mt{\hat{R}}\m{\bar{S}}_{i} = -\mt{\bar{D}}$ holds since $\m{\bar{S}}_{i}$ is obviously symmetric $\forall i$.

\end{document}

%% file: preamble_MMT.tex
\usepackage{siunitx}[=v2] 										
\usepackage{amsmath}
\usepackage{amsfonts}											
\usepackage{amssymb} 											
\usepackage{mathtools} 	 										
\usepackage{mathdots}											
\usepackage{bm,upgreek}											
\usepackage{relsize}											
\usepackage{xparse}												

\usepackage[ruled,noline]{algorithm2e}							
	\SetAlCapHSkip{0em}
	\DecMargin{2pt}
	\SetAlgoCaptionSeparator{}
	\DontPrintSemicolon
	\SetKwIF{If}{ElseIf}{Else}{if}{}{else if}{else}{end if}
	\SetKwFor{While}{while}{}{end while}
	\SetKwFor{For}{for}{}{end for}

\usepackage[utf8]{inputenc}										
\usepackage[british]{babel}										
\usepackage{datetime2}
	\DTMsetup{useregional}	 									
\usepackage{comment} 											
\usepackage{enumitem}											
\usepackage{appendix}											
\usepackage{hyperref} 											
\usepackage[nameinlink]{cleveref}								

\usepackage[dvipsnames]{xcolor} 	
\usepackage{soul}

\usepackage{float}	 											
\usepackage[position=bottom,
			font=footnotesize,
		 	skip=0pt]{subcaption}								
\usepackage{graphbox}											

\usepackage{booktabs}											
\usepackage{multirow} 											
\usepackage{threeparttable}										
\usepackage{array} 												
 	\newcommand{\PreserveBackslash}[1]{\let\temp=\\#1\let\\=\temp}
 	\newcolumntype{C}[1]{>{\PreserveBackslash\centering}p{#1}}
 	\newcolumntype{R}[1]{>{\PreserveBackslash\raggedleft}p{#1}}
 	\newcolumntype{L}[1]{>{\PreserveBackslash\raggedright}p{#1}}

\usepackage{setspace} 											
\usepackage{caption}											
\DeclareRobustCommand{\corr}[1]{
 	\textcolor{black}{
	\sethlcolor{GreenYellow}\hl{#1}}}

\newcommand\newl[2]{ 											
 	\expandafter\newlength\csname #1\endcsname
 	\expandafter\setlength\csname #1\endcsname{#2}}
\newcommand\setl[2]{ 											
 	\expandafter\setlength\csname #1\endcsname{#2}}

\newl{figh}{.0\columnwidth}										
\newl{figw}{.0\columnwidth}										
\newl{boxh}{.0\columnwidth}										
\newl{boxw}{.0\columnwidth}										
\newl{tcw}{.0\columnwidth}										

\newl{subfigskip}{.6em}											
\newl{plotvsep}{.8em}											

\usepackage{pgfplots}
\usepgfplotslibrary{groupplots}
\usetikzlibrary{positioning}

\usetikzlibrary{external}
\pgfplotsset{
	compat=newest,
	plot coordinates/math parser=false,
}
\pgfplotsset{
 	grid=both,
	minor grid style={black!5, ultra thin},
	major grid style={black!5, ultra thin},
	tick label style={font=\normalsize},
	label style={font=\normalsize},
 	legend style={font=\normalsize},
	legend cell align={left},
}
\pgfplotsset{
	every axis plot/.append style={semithick},
	lin subplots/.style={
		width=.85\columnwidth, 
		height=.41\columnwidth, 
		xminorticks=true,
		line join=round,
		},
	log subplots/.style={
		lin subplots,
		xmin=1, xmax=1000, xmode=log,
		ylabel={Magnitude (dB)}, 
		},
	matrix subplots/.style={
		width=.51\columnwidth,
		height=.31\columnwidth,
		xmin=0.1,xmax=1000,xmode=log,xminorticks=true,
		line join=round,
		tick label style={font=\small},
		grid=none,
		},
	matrix subplots lgd/.style={
		draw=none,
 		fill=lg!6,
 		/tikz/every even column/.append style={column sep=9pt},
 		legend image post style={scale=1.25},
		},
 	hexapod 1/.style={densely dotted},
	hexapod 2/.style={bc,thin},
	hexapod 3/.style={rc,thin},
}
\usetikzlibrary{spy}
\tikzset{rectagular spy/.style={%
	spy scope={%
		magnification=4,
	 	connect spies,
 	 	every spy on node/.style={rectangle,draw,},
 	 	every spy in node/.style={draw,rectangle,}
 		}
 	}
} 
\pgfdeclarelayer{annotation}
\pgfdeclarelayer{grid}
	\pgfsetlayers{grid,main,annotation}
		
\definecolor{rc}{rgb}{0.94,0.3,0.13}
\definecolor{bc}{rgb}{0.17,0.21,0.59}
\definecolor{dbc}{rgb}{0.10,0.12,0.35} 		
\definecolor{lg}{rgb}{0.30,0.40,0.45} 		
\definecolor{r2}{rgb}{0.8314,0.0118,0.0} 	

\DeclareSIUnit\dec{dec}											
\DeclareSIUnit\oct{oct}											

\DeclareMathSymbol{\shortminus}{\mathbin}{AMSa}{"39}			

\DeclareMathOperator{\diag}{diag}								

\newcommand{\invs}{{-1}}										
\newcommand{\pinvs}{+}											
\newcommand{\trs}{T}											

\newcommand{\skewt}[1]{{[#1]}_\times}							
\newcommand{\csys}[1]{\mathcal{#1}}						 		
\newcommand{\map}[1]{\mathcal{#1}}						 		

\newcommand{\inv}[1]{{#1}^\invs}							    
\newcommand{\pinv}[1]{{#1}^\pinvs}								
 \newcommand{\m}[1]{\mathbf{#1}}								
 \newcommand{\vc}[1]{\boldsymbol{\mathbf{#1}}}					
\newcommand{\tr}[1]{{#1}^\trs}									

\newcommand{\mi}[1]{\inv{\m{#1}}}								
\newcommand{\mpi}[1]{\pinv{\m{#1}}}								
\newcommand{\mt}[1]{\tr{\m{#1}}}								

\newcommand{\cA}{\cos\psi}
\newcommand{\cB}{\cos\theta}
\newcommand{\cC}{\cos\phi}
\newcommand{\sA}{\sin\psi}
\newcommand{\sB}{\sin\theta}
\newcommand{\sC}{\sin\phi}

\makeatletter
\NewDocumentCommand\getfontsize{m}{{\corr{Current font size:} \textbf{\f@size\ pt}}\par}
\makeatother

%% file: glossaries_MMT.tex
\usepackage[acronyms,nonumberlist,nopostdot,nomain]{glossaries} 
\makeglossaries 

\newacronym[]{acr:rhs}{RHS}{right-hand side}
\newacronym[]{acr:lhs}{LHS}{left-hand side}
\newacronym[]{acr:ode}{ODE}{ordinary differential equation}
\newacronym[]{acr:pde}{PDE}{partial differential equation}
\newacronym[]{acr:svd}{SVD}{singular value decomposition}
\newacronym[]{acr:bc}{BC}{boundary condition}
\newacronym[]{acr:psd}{PSD}{positive semi-definite}
\newacronym[]{acr:pd}{PD}{positive definite}
\newacronym[]{acr:gep}{GEP}{generalised eigenvalue problem}
\newacronym[]{acr:pdf}{PDF}{probability density function}
\newacronym[]{acr:gmes}{GMRES}{generalised minimum residual}
\newacronym[]{acr:fft}{FFT}{fast Fourier transform}
\newacronym[]{acr:dft}{DFT}{discrete Fourier transform}
\newacronym[longplural={coefficients of variation}]{acr:cov}{COV}{coefficient of variation}

\newacronym[longplural={degrees of freedom}]{acr:dof}{DOF}{degree of freedom}
\newacronym[longplural={natural frequencies}]{acr:nf}{NF}{natural frequency}
\newacronym[]{acr:ds}{DS}{dynamic substructuring}
\newacronym[]{acr:cms}{CMS}{component mode synthesis}
\newacronym[]{acr:spc}{SPC}{single point constraint}
\newacronym[]{acr:mpc}{MPC}{multi point constraint}
\newacronym[]{acr:mpf}{MPF}{modal participation factor}
\newacronym[longplural={non-structural masses}]{acr:nsm}{NSM}{non-structural mass}
\newacronym[]{acr:tf}{TF}{transfer function}

\newacronym[]{acr:fe}{FE}{finite element}
\newacronym[]{acr:fem}{FEM}{finite element method}
\newacronym[]{acr:cb}{CB}{Craig-Bampton}
\newacronym[]{acr:cbsm}{CBSM}{Craig-Bampton stochastic method}
\newacronym[]{acr:bem}{BEM}{boundary element method}
\newacronym[]{acr:sfem}{SFEM}{stochastic finite element method}
\newacronym[]{acr:sea}{SEA}{statistical energy analysis}
\newacronym[]{acr:psa}{PSA}{probabilistic structural analysis}
\newacronym[]{acr:mc}{MC}{Monte Carlo}
\newacronym[]{acr:mcs}{MCS}{Monte Carlo simulation}

\newacronym[]{acr:io}{I/O}{input/output}
\newacronym[longplural={centres of mass}]{acr:cm}{CM}{centre of mass}
\newacronym[longplural={centres of gravity}]{acr:cg}{CG}{centre of gravity}
\newacronym[longplural={equations of motion}]{acr:eqm}{EQM}{equtaion of motion}

\newacronym[]{acr:sstl}{SSTL}{Surrey Satellite Technology Limited}
\newacronym[]{acr:ssc}{SSC}{Surrey Space Centre}
\newacronym[]{acr:esa}{ESA}{European Space Agency}
\newacronym[]{acr:csem}{CSEM}{Centre Suisse d'Electronique et de Microtechnique}
\newacronym[]{acr:tpa-si}{TPA-SI}{Te P\={u}naha \={A}tea - Space Institute}

\newacronym[]{acr:rw}{RW}{reaction wheel}
\newacronym[longplural={reaction wheel assemblies}]{acr:rwa}{RWA}{reaction wheel assembly}
\newacronym[]{acr:emsd}{EMSD}{electromagnetic shunt damper}
\newacronym[]{acr:wedm}{WEDM}{wire electrical discharge machining}